\documentclass[twoside,11pt]{article}
\usepackage{jair}
\usepackage{theapa}
\usepackage{amsmath}
\makeatletter
\renewcommand{\jairheading}[5]{\def\ps@jairtps{\let\@mkboth\@gobbletwo%
\def\@oddhead{\scriptsize Journal of Artificial Intelligence Research #1 (#2) #3 \hfill Submitted #4; published #5}%
\def\@oddfoot{\scriptsize \copyright #2 AI Access Foundation and National Research
Council Canada. Reprinted with permission. \hfill}%
\def\@evenhead{}\def\@evenfoot{}}%
\thispagestyle{jairtps}}
\makeatother
\jairheading{37}{2010}{141-188}{10/09}{02/10}
\ShortHeadings{From Frequency to Meaning}{Turney \& Pantel}
\firstpageno{141}
\makeatletter
\newcommand\figcaption{\def\@captype{figure}\caption}
\newcommand\tabcaption{\def\@captype{table}\caption}
\makeatother
\newenvironment{myitemize}{
\vspace{-0.3\baselineskip}
\begin{itemize}
  \setlength{\topsep}{0pt}
  \setlength{\itemsep}{3pt}
  \setlength{\parskip}{0pt}
  \setlength{\parsep}{0pt}
  \setlength{\partopsep}{0pt}
}{
\end{itemize}
\vspace{-0.2\baselineskip}}
\newenvironment{myenumerate}{
\vspace{-0.3\baselineskip}
\begin{enumerate}
  \setlength{\topsep}{0pt}
  \setlength{\itemsep}{3pt}
  \setlength{\parskip}{0pt}
  \setlength{\parsep}{0pt}
  \setlength{\partopsep}{0pt}
}{
\end{enumerate}
\vspace{-0.2\baselineskip}}
\begin{document}

\title{From Frequency to Meaning: \\
       Vector Space Models of Semantics}

\author{\name Peter D. Turney \email peter.turney@nrc-cnrc.gc.ca \\
       \addr National Research Council Canada \\
       Ottawa, Ontario, Canada, K1A 0R6
       \AND
       \name Patrick Pantel \email me@patrickpantel.com \\
       \addr Yahoo! Labs \\
       Sunnyvale, CA, 94089, USA }

\maketitle

\begin{abstract}

Computers understand very little of the meaning of human language. This
profoundly limits our ability to give instructions to computers, the ability
of computers to explain their actions to us, and the ability of computers
to analyse and process text. Vector space models (VSMs) of semantics are
beginning to address these limits. This paper surveys the use of VSMs for
semantic processing of text. We organize the literature on VSMs
according to the structure of the matrix in a VSM. There are currently
three broad classes of VSMs, based on term--document, word--context, and
pair--pattern matrices, yielding three classes of applications. We
survey a broad range of applications in these three categories and
we take a detailed look at a specific open source project in each
category. Our goal in this survey is to show the breadth of applications
of VSMs for semantics, to provide a new perspective on VSMs for
those who are already familiar with the area, and to provide pointers
into the literature for those who are less familiar with the field.

\end{abstract}

\section{Introduction}
\label{sec:intro}

One of the biggest obstacles to making full use of the power of
computers is that they currently understand very little of the
meaning of human language. Recent progress in search engine technology
is only scratching the surface of human language, and yet the impact on
society and the economy is already immense. This hints at the transformative
impact that deeper semantic technologies will have. Vector space models
(VSMs), surveyed in this paper, are likely to be a part of these new semantic
technologies.

In this paper, we use the term {\em semantics} in a general sense, as the
meaning of a word, a phrase, a sentence, or any text in human language,
and the study of such meaning. We are not concerned with narrower senses
of {\em semantics}, such as the {\em semantic web} or approaches
to semantics based on formal logic. We present a survey of VSMs
and their relation with the distributional hypothesis
as an approach to representing some aspects of natural language semantics.

The VSM was developed for the SMART information retrieval
system \cite{salton71} by Gerard Salton and his colleagues \cite{salton75}.
SMART pioneered many of the concepts that are used in modern search engines
\cite{manning08}. The idea of the VSM is to represent each document in a
collection as a point in a space (a vector in a vector space). Points that
are close together in this space are semantically similar and points that are
far apart are semantically distant. The user's query is represented as a point in
the same space as the documents (the query is a {\em pseudo-document}).
The documents are sorted in order of increasing distance (decreasing semantic
similarity) from the query and then presented to the user.

The success of the VSM for information retrieval has inspired researchers to
extend the VSM to other semantic tasks in natural language processing, with
impressive results. For instance, \citeauthor{rapp03}~\citeyear{rapp03} used a
vector-based representation of word meaning to achieve a score
of 92.5\% on multiple-choice synonym questions from the Test
of English as a Foreign Language (TOEFL), whereas the average
human score was 64.5\%.\footnote{Regarding the average score of 64.5\% on the
TOEFL questions, \citeauthor{landauer97}~\citeyear{landauer97} note that, ``Although
we do not know how such a performance would compare, for example, with U.S. school
children of a particular age, we have been told that the average score is adequate
for admission to many universities.''} \citeauthor{turney06}~\citeyear{turney06} used a
vector-based representation of semantic relations to attain a score of 56\% on multiple-choice
analogy questions from the SAT college entrance test, compared to an average
human score of 57\%.\footnote{This is the average score for highschool students
in their senior year, applying to US universities. For more discussion of this
score, see Section 6.3 in \citeauthor{turney06}'s \citeyear{turney06} paper.}

In this survey, we have organized past work with VSMs according to the type
of matrix involved: term--document, word--context, and pair--pattern. We believe
that the choice of a particular matrix type is more fundamental than other
choices, such as the particular linguistic processing or mathematical processing.
Although these three matrix types cover most of the work, there is no reason
to believe that these three types exhaust the possibilities. We expect future
work will introduce new types of matrices and higher-order tensors.\footnote{A
vector is a first-order tensor and a matrix is a second-order tensor.
See Section~\ref{subsec:tensors}.}

\subsection{Motivation for Vector Space Models of Semantics}
\label{subsec:vector-motivation}

VSMs have several attractive properties. VSMs extract
knowledge automatically from a given corpus, thus they require
much less labour than other approaches to semantics, such as hand-coded
knowledge bases and ontologies. For example, the main resource used in
Rapp's~\citeyear{rapp03} VSM system for measuring word similarity is
the British National Corpus (BNC),\footnote{See http://www.natcorp.ox.ac.uk/.}
whereas the main resource used in non-VSM systems for measuring word similarity
\cite{hirst98,leacock98,jarmasz03} is a lexicon, such as
WordNet\footnote{See http://wordnet.princeton.edu/.} or Roget's Thesaurus.
Gathering a corpus for a new language is generally much easier than building
a lexicon, and building a lexicon often involves also gathering
a corpus, such as SemCor for WordNet \cite{miller93}.

VSMs perform well on tasks that involve measuring the similarity
of meaning between words, phrases, and documents. Most search engines
use VSMs to measure the similarity between a query and a document
\cite{manning08}. The leading algorithms for measuring semantic relatedness use
VSMs \cite{pantel02a,rapp03,turney03b}. The leading algorithms for measuring the
similarity of semantic relations also use VSMs \cite{lin01,turney06,nakov08}.
(Section~\ref{subsec:similarities} discusses the differences between these
types of similarity.)

We find VSMs especially interesting due to their relation
with the distributional hypothesis and related hypotheses (see
Section~\ref{subsec:hypotheses}). The distributional hypothesis is that
words that occur in similar contexts tend to have similar meanings
\cite{wittgenstein53,harris54,weaver55,firth57,deerwester90}. Efforts to
apply this abstract hypothesis to concrete algorithms for measuring the similarity
of meaning often lead to vectors, matrices, and higher-order tensors.
This intimate connection between the distributional hypothesis and
VSMs is a strong motivation for taking a close look at VSMs.

Not all uses of vectors and matrices count as vector space models.
For the purposes of this survey, we take it as a defining property
of VSMs that the values of the elements in a VSM must be derived from event
frequencies, such as the number of times that a given word appears in a given
context (see Section~\ref{subsec:type-token}). For example, often a lexicon or
a knowledge base may be viewed as a graph, and a graph may be represented using
an adjacency matrix, but this does not imply that a lexicon is a VSM, because,
in general, the values of the elements in an adjacency matrix are not derived
from event frequencies. This emphasis on event frequencies brings unity to the
variety of VSMs and explicitly connects them to the distributional hypothesis;
furthermore, it avoids triviality by excluding many possible matrix representations.

\subsection{Vectors in AI and Cognitive Science}
\label{subsec:ai}

Vectors are common in AI and cognitive science; they were common
before the VSM was introduced by \citeauthor{salton75}~\citeyear{salton75}.
The novelty of the VSM was to use frequencies in a corpus of text
as a clue for discovering semantic information.

In machine learning, a typical problem is to learn to classify
or cluster a set of items (i.e., examples, cases, individuals, entities)
represented as {\em feature vectors} \cite{mitchell97,witten05}. In
general, the features are not derived from event frequencies, although
this is possible (see Section~\ref{subsec:ml}). For example, a machine learning
algorithm can be applied to classifying or clustering documents \cite{sebastiani02}.

Collaborative filtering and recommender systems also use vectors
\cite{resnick94,breese98,linden03}. In a typical recommender system, we have
a {\em person-item} matrix, in which the rows correspond to people (customers,
consumers), the columns correspond to items (products, purchases), and
the value of an element is the rating (poor, fair, excellent) that the
person has given to the item. Many of the mathematical techniques that work
well with term--document matrices (see Section~\ref{sec:mathematical}) also
work well with person-item matrices, but ratings are not derived from event
frequencies.

In cognitive science, prototype theory often makes use of vectors.
The basic idea of prototype theory is that some members of a category
are more central than others \cite{rosch78,lackoff87}. For example,
{\em robin} is a central (prototypical) member of the category {\em bird},
whereas {\em penguin} is more peripheral. Concepts have varying degrees
of membership in categories ({\em graded} categorization). A natural
way to formalize this is to represent concepts as vectors and categories
as sets of vectors \cite{nosofsky86,smith88}. However, these vectors
are usually based on numerical scores that are elicited by
questioning human subjects; they are not based on event frequencies.

Another area of psychology that makes extensive use of vectors is
psychometrics, which studies the measurement of psychological abilities
and traits. The usual instrument for measurement is a test or
questionnaire, such as a personality test. The results of a test
are typically represented as a {\em subject-item} matrix, in which
the rows represent the subjects (people) in an experiment and the columns
represent the items (questions) in the test (questionnaire). The value
of an element in the matrix is the answer that the corresponding
subject gave for the corresponding item. Many techniques for vector analysis,
such as factor analysis \cite{spearman04}, were pioneered in psychometrics.

In cognitive science, Latent Semantic Analysis (LSA) \cite{deerwester90,landauer97},
Hyperspace Analogue to Language (HAL) \cite{lund95,lund96}, and related
research \cite{landauer07} is entirely within the scope of VSMs, as
defined above, since this research uses vector space models in which the
values of the elements are derived from event frequencies, such as the number
of times that a given word appears in a given context. Cognitive scientists
have argued that there are empirical and theoretical reasons for believing
that VSMs, such as LSA and HAL, are plausible models of some aspects
of human cognition \cite{landauer07}. In AI, computational linguistics, and
information retrieval, such plausibility is not essential, but it may be seen
as a sign that VSMs are a promising area for further research.

\subsection{Motivation for This Survey}
\label{subsec:survey-motivation}

This paper is a survey of vector space models of semantics. There is
currently no comprehensive, up-to-date survey of this field. As we show
in the survey, vector space models are a highly successful approach to
semantics, with a wide range of potential and actual applications. There
has been much recent growth in research in this area.

This paper should be of interest to all AI researchers who work with natural language,
especially researchers who are interested in semantics. The survey will serve as a
general introduction to this area and it will provide a framework
-- a unified perspective -- for organizing the diverse literature on the
topic. It should encourage new research in the area, by pointing out open
problems and areas for further exploration.

This survey makes the following contributions:

{\bf New framework:} We provide a new framework for organizing the literature:
term--document, word--context, and pair--pattern matrices (see Section~\ref{sec:vsms}).
This framework shows the importance of the structure of the matrix (the choice of
rows and columns) in determining the potential applications and may inspire
researchers to explore new structures (different kinds of rows and columns, or
higher-order tensors instead of matrices).

{\bf New developments:} We draw attention to pair--pattern matrices. The use of
pair--pattern matrices is relatively new and deserves more study. These matrices
address some criticisms that have been directed at word--context matrices,
regarding lack of sensitivity to word order.

{\bf Breadth of approaches and applications:} There is no existing survey
that shows the breadth of potential and actual applications of VSMs for semantics.
Existing summaries omit pair--pattern matrices \cite{landauer07}.

{\bf Focus on NLP and CL:} Our focus in this survey is on systems that
perform practical tasks in natural language processing and computational
linguistics. Existing overviews focus on cognitive psychology \cite{landauer07}.

{\bf Success stories:} We draw attention to the fact that VSMs are arguably
the most successful approach to semantics, so far.

\subsection{Intended Readership}
\label{subsec:readership}

Our goal in writing this paper has been to survey the state of the
art in vector space models of semantics, to introduce the topic to
those who are new to the area, and to give a new perspective to those
who are already familiar with the area.

We assume our reader has a basic understanding of vectors, matrices, and
linear algebra, such as one might acquire from an introductory undergraduate
course in linear algebra, or from a text book \cite{golub96}.
The basic concepts of vectors and matrices are more important here than the
mathematical details. \citeauthor{widdows04}~\citeyear{widdows04} gives
a gentle introduction to vectors from the perspective of semantics.

We also assume our reader has some familiarity with computational
linguistics or information retrieval. \citeauthor{manning08}~\citeyear{manning08}
provide a good introduction to information retrieval.
For computational linguistics, we recommend \citeauthor{manning99}'s
\citeyear{manning99} text.

If our reader is familiar with linear algebra and computational
linguistics, this survey should present no barriers to understanding.
Beyond this background, it is not necessary to be familiar with
VSMs as they are used in information retrieval, natural language
processing, and computational linguistics. However, if the reader
would like to do some further background reading, we recommend
\citeauthor{landauer07}'s \citeyear{landauer07} collection.

\subsection{Highlights and Outline}
\label{subsec:outline}

This article is structured as follows. Section~\ref{sec:vsms} explains
our framework for organizing the literature on VSMs according to the type
of matrix involved: term--document, word--context, and pair--pattern. In this
section, we present an overview of VSMs, without getting into the details
of how a matrix can be generated from a corpus of raw text.

After the high-level framework is in place, Sections \ref{sec:linguistic}
and \ref{sec:mathematical} examine the steps involved in generating
a matrix. Section~\ref{sec:linguistic} discusses linguistic
processing and Section~\ref{sec:mathematical} reviews mathematical
processing. This is the order in which a corpus would be processed
in most VSM systems (first linguistic processing, then mathematical
processing).

When VSMs are used for semantics, the input to the model is
usually plain text. Some VSMs work directly with the raw text,
but most first apply some linguistic processing to the text,
such as stemming, part-of-speech tagging, word sense tagging, or
parsing. Section~\ref{sec:linguistic} looks at some of these
linguistic tools for semantic VSMs.

In a simple VSM, such as a simple term--document VSM,
the value of an element in a document vector
is the number of times that the corresponding word occurs
in the given document, but most VSMs apply some mathematical processing
to the raw frequency values. Section~\ref{sec:mathematical} presents
the main mathematical operations: weighting the elements, smoothing
the matrix, and comparing the vectors. This section also
describes optimization strategies for comparing the vectors, such as
distributed sparse matrix multiplication and randomized techniques.

By the end of Section~\ref{sec:mathematical}, the reader will have a
general view of the concepts involved in vector space models of semantics.
We then take a detailed look at three VSM systems in Section~\ref{sec:three-systems}.
As a representative of term--document VSMs, we present the Lucene information
retrieval library.\footnote{See http://lucene.apache.org/java/docs/.}
For word--context VSMs, we explore the Semantic Vectors package, which builds on
Lucene.\footnote{See http://code.google.com/p/semanticvectors/.}
As the representative of pair--pattern VSMs, we review the Latent Relational
Analysis module in the S-Space package, which also builds on
Lucene.\footnote{See http://code.google.com/p/airhead-research/wiki/LatentRelationalAnalysis.}
The source code for all three of these systems is available under
open source licensing.

We turn to a broad survey of applications for semantic VSMs in
Section~\ref{sec:applications}. This section also
serves as a short historical view of research with semantic VSMs, beginning
with information retrieval in Section~\ref{subsec:term--document-apps}.
Our purpose here is to give the reader an idea of the breadth of applications
for VSMs and also to provide pointers into the literature, if the reader
wishes to examine any of these applications in detail.

In a {\em term--document matrix}, rows correspond to terms and columns correspond
to documents (Section~\ref{subsec:term--document-apps}). A document provides a
context for understanding the term. If we generalize the idea of documents to chunks
of text of arbitrary size (phrases, sentences, paragraphs, chapters, books, collections),
the result is the {\em word--context} matrix, which includes the term--document
matrix as a special case. Section~\ref{subsec:word--context-apps} discusses
applications for word--context matrices. Section~\ref{subsec:pair--pattern-apps}
considers {\em pair--pattern} matrices, in which the rows correspond to
pairs of terms and the columns correspond to the patterns in which
the pairs occur.

In Section~\ref{sec:alternatives}, we discuss alternatives to VSMs for semantics.
Section~\ref{sec:future} considers the future of VSMs, raising some questions
about their power and their limitations. We conclude in Section~\ref{sec:conclusions}.

\section{Vector Space Models of Semantics}
\label{sec:vsms}

The theme that unites the various forms of VSMs that we discuss in this paper can be
stated as the {\em statistical semantics hypothesis}: statistical patterns of human word
usage can be used to figure out what people mean.\footnote{This phrase was
taken from the Faculty Profile of George Furnas at the University of Michigan,
http://www.si.umich.edu/people/faculty-detail.htm?sid=41. The full quote
is, ``Statistical Semantics -- Studies of how the statistical patterns of human
word usage can be used to figure out what people mean, at least to a level sufficient
for information access.'' The term {\em statistical semantics} appeared in the work of
\citeauthor{furnas83}~\citeyear{furnas83}, but it was not defined there.} This general
hypothesis underlies several more specific hypotheses, such as the {\em bag of words
hypothesis}, the {\em distributional hypothesis}, the {\em extended distributional
hypothesis}, and the {\em latent relation hypothesis}, discussed below.

\subsection{Similarity of Documents: The Term--Document Matrix}
\label{subsec:term--document-vsms}

In this paper, we use the following notational conventions:
Matrices are denoted by bold capital letters, $\mathbf{A}$.
Vectors are denoted by bold lowercase letters, $\mathbf{b}$. Scalars are
represented by lowercase italic letters, $c$.

If we have a large collection of documents, and hence a large number of document
vectors, it is convenient to organize the vectors into a matrix. The row vectors
of the matrix correspond to terms (usually terms are words, but we will discuss
some other possibilities) and the column vectors correspond to documents (web pages,
for example). This kind of matrix is called a {\em term--document} matrix.

In mathematics, a {\em bag} (also called a {\em multiset}) is like a set, except
that duplicates are allowed. For example, $\{ a, a, b, c, c, c \}$ is a bag
containing $a$, $b$, and $c$. Order does not matter in bags and sets; the
bags $\{ a, a, b, c, c, c \}$ and  $\{ c, a, c, b, a, c \}$ are equivalent.
We can represent the bag $\{ a, a, b, c, c, c \}$ with the vector
$\mathbf{x} = \langle 2, 1, 3 \rangle$, by stipulating that the first
element of $\mathbf{x}$ is the frequency of $a$ in the bag, the second element
is the frequency of $b$ in the bag, and the third element is the frequency of $c$.
A set of bags can be represented as a matrix $\mathbf{X}$, in which each column
$\mathbf{x}_{:j}$ corresponds to a bag, each row $\mathbf{x}_{i:}$ corresponds to
a unique member, and an element $x_{ij}$  is the frequency of the $i$-th
member in the $j$-th bag.

In a term--document matrix, a document vector represents the corresponding document
as a bag of words. In information retrieval, the {\em bag of words hypothesis} is
that we can estimate the relevance of documents to a query by representing the
documents and the query as bags of words. That is, the frequencies of words in a
document tend to indicate the relevance of the document to a query. The bag of
words hypothesis is the basis for applying the VSM to information retrieval
\cite{salton75}. The hypothesis expresses the belief that a column vector in a
term--document matrix captures (to some degree) an aspect of the meaning of the
corresponding document; what the document is {\em about}.

Let $\mathbf{X}$ be a term--document matrix. Suppose our document
collection contains $n$ documents and $m$ unique terms. The matrix $\mathbf{X}$
will then have $m$ rows (one row for each unique term in the vocabulary)
and $n$ columns (one column for each document). Let $w_i$ be the
$i$-th term in the vocabulary and let $d_j$ be the $j$-th
document in the collection. The $i$-th row in $\mathbf{X}$ is the row vector
$\mathbf{x}_{i:}$ and the $j$-th column in $\mathbf{X}$ is
the column vector $\mathbf{x}_{:j}$. The row vector $\mathbf{x}_{i:}$
contains $n$ elements, one element for each document, and the column
vector $\mathbf{x}_{:j}$ contains $m$ elements, one element for each term.
Suppose $\mathbf{X}$ is a simple matrix of frequencies.
The element $x_{ij}$ in $\mathbf{X}$ is the frequency of the $i$-th
term $w_i$ in the $j$-th document $d_j$.

In general, the value of most of the elements in $\mathbf{X}$ will be
zero (the matrix is {\em sparse}), since most documents will use
only a small fraction of the whole vocabulary. If we randomly choose
a term $w_i$ and a document $d_j$, it's likely that $w_i$
does not occur anywhere in $d_j$, and therefore $x_{ij}$ equals 0.

The pattern of numbers in $\mathbf{x}_{i:}$ is
a kind of {\em signature} of the $i$-th term $w_i$; likewise,
the pattern of numbers in $\mathbf{x}_{:j}$ is a signature
of the $j$-th document $d_j$. That is, the pattern of numbers
tells us, to some degree, what the term or document is {\em about}.

The vector $\mathbf{x}_{:j}$ may seem to be a rather crude representation
of the document $d_j$. It tells us how frequently the words appear in the
document, but the sequential order of the words is lost. The vector
does not attempt to capture the structure in the phrases, sentences,
paragraphs, and chapters of the document. However, in spite of this
crudeness, search engines work surprisingly well; vectors seem to
capture an important aspect of semantics.

The VSM of \citeauthor{salton75}~\citeyear{salton75} was arguably the first
practical, useful algorithm for extracting semantic information from word usage.
An intuitive justification for the term--document matrix is that the topic of a
document will probabilistically influence the author's choice of words when
writing the document.\footnote{Newer generative models, such as Latent Dirichlet
Allocation (LDA) \cite{blei03}, directly model this intuition. See
Sections \ref{subsec:smoothing} and \ref{sec:alternatives}.} If two documents
have similar topics, then the two corresponding column vectors will tend to
have similar patterns of numbers.

\subsection{Similarity of Words: The Word--Context Matrix}
\label{subsec:word--context-vsms}

\citeauthor{salton75}~\citeyear{salton75} focused on measuring document similarity,
treating a query to a search engine as a pseudo-document. The relevance
of a document to a query is given by the similarity of their vectors.
\citeauthor{deerwester90}~\citeyear{deerwester90} observed that we can shift the
focus to measuring word similarity, instead of document similarity, by looking
at row vectors in the term--document matrix, instead of column vectors.

\citeauthor{deerwester90}~\citeyear{deerwester90} were inspired by
the term--document matrix of \citeauthor{salton75}~\citeyear{salton75},
but a document is not necessarily the optimal length of text for measuring
word similarity. In general, we may have a {\em word--context} matrix, in which the
context is given by words, phrases, sentences, paragraphs, chapters, documents,
or more exotic possibilities, such as sequences of characters or patterns.

The {\em distributional hypothesis} in linguistics is that words that
occur in similar contexts tend to have similar meanings \cite{harris54}.
This hypothesis is the justification for applying the VSM to
measuring word similarity. A word may be represented by a vector in which
the elements are derived from the occurrences of the word in various contexts,
such as windows of words \cite{lund96}, grammatical dependencies \cite{lin98b,pado07},
and richer contexts consisting of dependency links and selectional preferences on
the argument positions \cite{erk08}; see \citeauthor{sahlgren06}'s \citeyear{sahlgren06}
thesis for a comprehensive study of various contexts. Similar row vectors in the
word--context matrix indicate similar word meanings.

The idea that word usage can reveal semantics was implicit in some of the
things that \citeauthor{wittgenstein53}~\citeyear{wittgenstein53} said about language-games
and family resemblance. Wittgenstein was primarily interested in the physical
activities that form the context of word usage (e.g., the word {\em brick},
spoken in the context of the physical activity of building a house), but the
main context for a word is often other words.\footnote{Wittgenstein's
intuition might be better captured by a matrix that combines words
with other modalities, such as images \cite{monay03}. If the values of the
elements are derived from event frequencies, we would include this as a VSM
approach to semantics.}

\citeauthor{weaver55}~\citeyear{weaver55} argued that word sense disambiguation for machine
translation should be based on the co-occurrence frequency of the context words
near a given target word (the word that we want to disambiguate). \citeauthor{firth57}
\citeyear[p.~11]{firth57} said, ``You shall know a word by the company it keeps.''
\citeauthor{deerwester90}~\citeyear{deerwester90} showed how the
intuitions of \citeauthor{wittgenstein53}~\citeyear{wittgenstein53}, \citeauthor{harris54}
\citeyear{harris54}, \citeauthor{weaver55}, and \citeauthor{firth57}
could be used in a practical algorithm.

\subsection{Similarity of Relations: The Pair--Pattern Matrix}
\label{subsec:pair--pattern-vsms}

In a {\em pair--pattern} matrix, row vectors correspond to pairs of words,
such as {\em mason}$\,:\,${\em stone} and {\em carpenter}$\,:\,${\em wood}, and
column vectors correspond to the patterns in which the pairs
co-occur, such as ``$X$ cuts $Y$'' and ``$X$ works with $Y$''. \citeauthor{lin01}
\citeyear{lin01} introduced the pair--pattern matrix for the purpose of measuring
the semantic similarity of patterns; that is, the similarity of column vectors.
Given a pattern such as ``$X$ solves $Y$'', their algorithm was able to find
similar patterns, such as ``$Y$ is solved by $X$'', ``$Y$ is resolved in $X$'',
and ``$X$ resolves $Y$''.

\citeauthor{lin01}~\citeyear{lin01} proposed the {\em extended distributional
hypothesis}, that patterns that co-occur with similar pairs tend to have
similar meanings. The patterns ``$X$ solves $Y$'' and ``$Y$ is solved by $X$'' tend
to co-occur with similar $X\!:Y$ pairs, which suggests that these patterns have
similar meanings. Pattern similarity can be used to infer that one sentence
is a paraphrase of another \cite{lin01}.

\citeauthor{turney03b}~\citeyear{turney03b} introduced the use of the pair--pattern
matrix for measuring the semantic similarity of relations between word pairs;
that is, the similarity of row vectors. For example, the pairs
{\em mason}$\,:\,${\em stone}, {\em carpenter}$\,:\,${\em wood},
{\em potter}$\,:\,${\em clay}, and {\em glass\-blower}$\,:\,${\em glass} share
the semantic relation {\em artisan}$\,:\,${\em material}. In each case, the first
member of the pair is an artisan who makes artifacts from the material that is the
second member of the pair. The pairs tend to co-occur in similar patterns, such as
``the $X$ used the $Y$ to'' and ``the $X$ shaped the $Y$ into''.

The {\em latent relation hypothesis} is that pairs of words that co-occur
in similar patterns tend to have similar semantic relations \cite{turney08b}.
Word pairs with similar row vectors in a pair--pattern matrix tend to have similar
semantic relations. This is the inverse of the extended distributional hypothesis,
that patterns with similar column vectors in the pair--pattern matrix
tend to have similar meanings.

\subsection{Similarities}
\label{subsec:similarities}

Pair--pattern matrices are suited to measuring the similarity of semantic relations
between pairs of words; that is, {\em relational similarity}. In contrast,
word--context matrices are suited to measuring {\em attributional similarity}.
The distinction between attributional and relational similarity has
been explored in depth by \citeauthor{gentner83}~\citeyear{gentner83}.

The attributional similarity between two words $a$ and $b$, ${\rm sim_a}(a,b) \in \Re$,
depends on the degree of correspondence between the properties of $a$ and $b$.
The more correspondence there is, the greater their attributional similarity.
The relational similarity between two {\em pairs} of words $a\!:\!b$ and $c\!:\!d$,
${\rm sim_r}(a\!:\!b, c\!:\!d) \in \Re$, depends on the degree of correspondence
between the relations of $a\!:\!b$ and $c\!:\!d$. The more correspondence there is,
the greater their relational similarity. For example, {\em dog} and {\em wolf}
have a relatively high degree of attributional similarity, whereas
{\em dog}$\,:\,${\em bark} and {\em cat}$\,:\,${\em meow} have a relatively
high degree of relational similarity \cite{turney06}.

It is tempting to suppose that relational similarity can be reduced
to attributional similarity. For example, {\em mason} and {\em carpenter}
are similar words and {\em stone} and {\em wood} are similar
words; therefore, perhaps it follows that {\em mason}$\,:\,${\em stone}
and {\em carpenter}$\,:\,${\em wood} have similar relations.
Perhaps ${\rm sim_r}(a\!:\!b, c\!:\!d)$ can be reduced
to ${\rm sim_a}(a,c) + {\rm sim_a}(b,d)$. However, {\em mason},
{\em carpenter}, {\em potter}, and
{\em glass\-blower} are similar words (they are all artisans), as are
{\em wood}, {\em clay}, {\em stone}, and {\em glass} (they are all
materials used by artisans), but we cannot infer from this
that {\em mason}$\,:\,${\em glass} and {\em carpenter}$\,:\,${\em clay}
have similar relations. Turney~\citeyear{turney06,turney08b} presented
experimental evidence that relational similarity does not reduce
to attributional similarity.

The term {\em semantic relatedness} in computational linguistics \cite{budanitsky01}
corresponds to attributional similarity in cognitive science \cite{gentner83}. Two
words are semantically related if they have any kind of semantic relation
\cite{budanitsky01}; they are semantically related to the degree that they
share attributes \cite{turney06}. Examples are synonyms ({\em bank} and
{\em trust company}), meronyms ({\em car} and {\em wheel}), antonyms
({\em hot} and {\em cold}), and words that are functionally related
or frequently associated ({\em pencil} and {\em paper}). We might not
usually think that antonyms are similar, but antonyms have a high degree of
attributional similarity (hot and cold are kinds of temperature,
black and white are kinds of colour, loud and quiet are kinds of sound).
We prefer the term {\em attributional similarity} to the term
{\em semantic relatedness}, because {\em attributional similarity} emphasizes
the contrast with {\em relational similarity}, whereas {\em semantic relatedness}
could be confused with {\em relational similarity}.

In computational linguistics, the term {\em semantic similarity} is applied
to words that share a hypernym ({\em car} and {\em bicycle} are semantically
similar, because they share the hypernym {\em vehicle}) \cite{resnik95}.
Semantic similarity is a specific type of attributional similarity.
We prefer the term {\em taxonomical similarity} to the term
{\em semantic similarity}, because the term {\em semantic similarity}
is misleading. Intuitively, both attributional and relational
similarity involve meaning, so both deserve to be called semantic
similarity.

Words are {\em semantically associated} if they tend to co-occur frequently
(e.g., {\em bee} and {\em honey}) \cite{chiarello90}.
Words may be taxonomically
similar and semantically associated ({\em doctor} and {\em nurse}),
taxonomically similar but not semantically associated ({\em horse}
and {\em platypus}), semantically associated but not taxonomically
similar ({\em cradle} and {\em baby}), or neither semantically
associated nor taxonomically similar ({\em calculus} and
{\em candy}).

\citeauthor{schutze93}~\citeyear{schutze93} defined two ways that
words can be distributed in a corpus of text: If two words tend to be neighbours
of each other, then they are {\em syntagmatic associates}. If two words
have similar neighbours, then they are {\em paradigmatic parallels}.
Syntagmatic associates are often different parts of speech, whereas
paradigmatic parallels are usually the same part of speech. Syntagmatic
associates tend to be semantically associated ({\em bee} and {\em honey}
are often neighbours); paradigmatic parallels tend to be taxonomically similar
({\em doctor} and {\em nurse} have similar neighbours).

\subsection{Other Semantic VSMs}
\label{subsec:tensors}

The possibilities are not exhausted by term--document, word--context, and
pair--pattern matrices. We might want to consider triple--pattern matrices, for
measuring the semantic similarity between word triples.
Whereas a pair--pattern matrix might have a row {\em mason}$\,:\,${\em stone}
and a column ``$X$ works with $Y$'', a triple--pattern matrix could have
a row {\em mason}$\,:\,${\em stone}$\,:\,${\em masonry} and a column
``$X$ uses $Y$ to build $Z$''.
However, $n$-tuples of words grow increasingly rare as $n$ increases.
For example, phrases that contain {\em mason}, {\em stone}, and {\em masonry}
together are less frequent than phrases that contain {\em mason} and {\em stone}
together. A triple--pattern matrix will be much more sparse than a
pair--pattern matrix (ceteris paribus).
The quantity of text that we need, in order to have enough numbers
to make our matrices useful, grows rapidly as $n$ increases. It
may be better to break $n$-tuples into pairs. For example,
$a\!:\!b\!:\!c$ could be decomposed into $a\!:\!b$, $a\!:\!c$, and $b\!:\!c$
\cite{turney08b}.
The similarity of two triples, $a\!:\!b\!:\!c$ and $d\!:\!e\!:\!f$,
could be estimated by the similarity of their corresponding pairs.
A relatively dense pair--pattern matrix could serve as a surrogate
for a relatively sparse triple--pattern matrix.

We may also go beyond matrices. The generalization of a matrix is a
tensor \cite{kolda09,acar09}. A scalar (a single number) is zeroth-order
tensor, a vector is first-order tensor, and a matrix is a second-order tensor.
A tensor of order three or higher is called a higher-order tensor.
\citeauthor{chew07}~\citeyear{chew07} use a term--document--language
third-order tensor for multilingual information retrieval.
\citeauthor{turney07}~\citeyear{turney07} uses a word--word--pattern
tensor to measure similarity of words. \citeauthor{vandecruys09}~\citeyear{vandecruys09}
uses a verb--subject--object tensor to learn selectional preferences
of verbs.

In \citeauthor{turney07}'s \citeyear{turney07} tensor, for example, rows correspond to
words from the TOEFL multiple-choice synonym questions, columns correspond
to words from Basic English \cite{ogden30},\footnote{Basic English
is a highly reduced subset of English, designed to be easy for people to learn. The
words of Basic English are listed at http://ogden.basic-english.org/.}
and {\em tubes} correspond
to patterns that join rows and columns (hence we have a word--word--pattern
third-order tensor). A given word from the TOEFL questions is represented
by the corresponding word--pattern matrix {\em slice} in the tensor.
The elements in this slice correspond to all the patterns that relate
the given TOEFL word to any word in Basic English. The similarity of
two TOEFL words is calculated by comparing the two corresponding matrix
slices. The algorithm achieves 83.8\% on the TOEFL questions.

\subsection{Types and Tokens}
\label{subsec:type-token}

A {\em token} is a single instance of a symbol, whereas a {\em type} is a
general class of tokens \cite{manning08}. Consider the following example
(from Samuel Beckett):

\begin{quote}
Ever tried. Ever failed. \\
No matter. Try again. \\
Fail again. Fail better.
\end{quote}

\noindent There are two tokens of the type {\em Ever}, two tokens of the type
{\em again}, and two tokens of the type {\em Fail}. Let's say that each line
in this example is a document, so we have three documents of two sentences each.
We can represent this example with a token--document matrix or a type--document
matrix. The token--document matrix has twelve rows, one for each token, and three
columns, one for each line (Figure~\ref{fig:token}). The type--document matrix
has nine rows, one for each type, and three columns (Figure~\ref{fig:type}).

\newcommand{\vl}{\vline \hspace{6pt}}
\newcommand{\vr}{\hspace{1pt} \vline}

\begin{table}[htbp]
\small
\centering
\begin{tabular}{lccc}
          & Ever tried.     & No matter.   & Fail again.     \\
          & Ever failed.    & Try again.   & Fail better.    \\
\\
Ever      &  \vl   1        &     0        &      0   \vr    \\
tried     &  \vl   1        &     0        &      0   \vr    \\
Ever      &  \vl   1        &     0        &      0   \vr    \\
failed    &  \vl   1        &     0        &      0   \vr    \\
No        &  \vl   0        &     1        &      0   \vr    \\
matter    &  \vl   0        &     1        &      0   \vr    \\
Try       &  \vl   0        &     1        &      0   \vr    \\
again     &  \vl   0        &     1        &      0   \vr    \\
Fail      &  \vl   0        &     0        &      1   \vr    \\
again     &  \vl   0        &     0        &      1   \vr    \\
Fail      &  \vl   0        &     0        &      1   \vr    \\
better    &  \vl   0        &     0        &      1   \vr    \\
\\
\end{tabular}
\normalsize
\figcaption {The token--document matrix. Rows are tokens and columns
are documents.}
\label{fig:token}
\end{table}

\begin{table}[htbp]
\small
\centering
\begin{tabular}{lccc}
          & Ever tried.     & No matter.   & Fail again.     \\
          & Ever failed.    & Try again.   & Fail better.    \\
\\
Ever      &  \vl   2        &     0        &      0   \vr    \\
tried     &  \vl   1        &     0        &      0   \vr    \\
failed    &  \vl   1        &     0        &      0   \vr    \\
No        &  \vl   0        &     1        &      0   \vr    \\
matter    &  \vl   0        &     1        &      0   \vr    \\
Try       &  \vl   0        &     1        &      0   \vr    \\
again     &  \vl   0        &     1        &      1   \vr    \\
Fail      &  \vl   0        &     0        &      2   \vr    \\
better    &  \vl   0        &     0        &      1   \vr    \\
\\
\end{tabular}
\normalsize
\figcaption {The type--document matrix. Rows are types and columns
are documents.}
\label{fig:type}
\end{table}

A row vector for a token has binary values: an element is 1 if the given token
appears in the given document and 0 otherwise. A row vector for a type has
integer values: an element is the frequency of the given type in the given document.
These vectors are related, in that a type vector is the sum of the corresponding
token vectors. For example, the row vector for the type {\em Ever} is the
sum of the two token vectors for the two tokens of {\em Ever}.

In applications dealing with polysemy, one approach uses vectors
that represent word tokens \cite{schutze98,aggire06} and another uses vectors
that represent word types \cite{pantel02a}.
Typical word sense disambiguation (WSD) algorithms deal with word tokens (instances
of words in specific contexts) rather than word types.
We mention both approaches to polysemy in
Section~\ref{sec:applications}, due to their similarity and close relationship,
although a defining characteristic of the VSM is that it is concerned with frequencies
(see Section~\ref{subsec:vector-motivation}), and frequency is a property of
types, not tokens.

\subsection{Hypotheses}
\label{subsec:hypotheses}

We have mentioned five hypotheses in this section. Here we repeat these
hypotheses and then interpret them in terms of vectors. For each hypothesis,
we cite work that explicitly states something like the hypothesis
or implicitly assumes something like the hypothesis.

{\bf Statistical semantics hypothesis:} Statistical patterns of human word
usage can be used to figure out what people mean \cite{weaver55,furnas83}.
-- If units of text have similar vectors in a text frequency
matrix,\footnote{By {\em text frequency matrix}, we
mean any matrix or higher-order tensor in which the values of the elements
are derived from the frequencies of pieces of text in the context of other
pieces of text in some collection of text. A text frequency matrix is
intended to be a general structure, which includes term--document,
word--context, and pair--pattern matrices as special cases.}
then they tend to have similar meanings. (We take this to be a general hypothesis
that subsumes the four more specific hypotheses that follow.)

{\bf Bag of words hypothesis:} The frequencies of words in a document tend
to indicate the relevance of the document to a query \cite{salton75}.
-- If documents and pseudo-documents (queries) have similar column vectors
in a term--document matrix, then they tend to have similar meanings.

{\bf Distributional hypothesis:} Words that occur in similar contexts
tend to have similar meanings \cite{harris54,firth57,deerwester90}. -- If
words have similar row vectors in a word--context matrix, then they tend to
have similar meanings.

{\bf Extended distributional hypothesis:} Patterns that co-occur with similar
pairs tend to have similar meanings \cite{lin01}. -- If patterns have similar column
vectors in a pair--pattern matrix, then they tend to express similar semantic
relations.

{\bf Latent relation hypothesis:} Pairs of words that co-occur in similar
patterns tend to have similar semantic relations \cite{turney03b}. -- If word
pairs have similar row vectors in a pair--pattern matrix, then they tend to have
similar semantic relations.

We have not yet explained what it means to say that vectors are similar. We
discuss this in Section~\ref{subsec:comparing}.

\section{Linguistic Processing for Vector Space Models}
\label{sec:linguistic}

We will assume that our raw data is a large corpus of natural language text.
Before we generate a term--document, word--context, or pair--pattern matrix, it
can be useful to apply some linguistic processing to the raw text.
The types of processing that are used can be grouped into three classes.
First, we need to {\em tokenize} the raw text; that is, we need to decide
what constitutes a {\em term} and how to extract terms from raw text.
Second, we may want to {\em normalize} the raw text, to convert superficially
different strings of characters to the same form (e.g., {\em car}, {\em Car},
{\em cars}, and {\em Cars} could all be normalized to {\em car}).
Third, we may want to {\em annotate} the raw text, to mark identical
strings of characters as being different (e.g., {\em fly} as a
verb could be annotated as {\em fly/VB} and {\em fly} as a noun
could be annotated as {\em fly/NN}).

\citeauthor{grefenstette94}~\citeyear{grefenstette94} presents a good study of
linguistic processing for word--context VSMs. He uses a similar three-step
decomposition of linguistic processing: tokenization, surface syntactic
analysis, and syntactic attribute extraction.

\subsection{Tokenization}

Tokenization of English seems simple at first glance: words are separated
by spaces. This assumption is approximately true for English, and it may work
sufficiently well for a basic VSM, but a more advanced VSM requires a more
sophisticated approach to tokenization.

An accurate English tokenizer must know how to handle punctuation
(e.g., {\em don't}, {\em Jane's}, {\em and/or}), hyphenation (e.g.,
{\em state-of-the-art} versus {\em state of the art}), and recognize
multi-word terms (e.g., {\em Barack Obama} and {\em ice hockey}) \cite{manning08}.
We may also wish to ignore {\em stop words}, high-frequency words with
relatively low information content, such as function words (e.g., {\em of},
{\em the}, {\em and}) and pronouns (e.g., {\em them}, {\em who}, {\em that}).
A popular list of stop words is the set of 571 common words included
in the source code for the SMART system \cite{salton71}.\footnote{The
source code is available at ftp://ftp.cs.cornell.edu/pub/smart/.}

In some languages (e.g., Chinese), words are not separated by spaces.
A basic VSM can break the text into character unigrams or bigrams.
A more sophisticated approach is to match the input text against entries
in a lexicon, but the matching often does not determine a unique tokenization
\cite{sproat03}. Furthermore, native speakers often disagree about the correct
segmentation. Highly accurate tokenization is a challenging task
for most human languages.

\subsection{Normalization}
\label{subsec:normalization}

The motivation for normalization is the observation that many different
strings of characters often convey essentially identical meanings.
Given that we want to get at the meaning that underlies the words, it
seems reasonable to normalize superficial variations by converting them
to the same form. The most common types of normalization are case folding
(converting all words to lower case) and stemming (reducing inflected words
to their stem or root form).

Case folding is easy in English, but can be problematic in some
languages. In French, accents are optional for uppercase,
and it may be difficult to restore missing accents when converting the
words to lowercase. Some words cannot be distinguished without
accents; for example, {\em PECHE} could be either {\em p{\^e}che}
(meaning {\em fishing} or {\em peach}) or {\em p{\'e}ch{\'e}}
(meaning {\em sin}). Even in English, case folding can cause problems,
because case sometimes has semantic significance. For example, {\em SMART}
is an information retrieval system, whereas {\em smart} is a common
adjective; {\em Bush} may be a surname, whereas {\em bush} is a kind
of plant.

Morphology is the study of the internal structure of words. Often a
word is composed of a stem (root) with added affixes (inflections),
such as plural forms and past tenses (e.g., {\em trapped} is composed
of the stem {\em trap} and the affix {\em -ed}). Stemming, a kind of
morphological analysis, is the process of reducing inflected words to
their stems. In English, affixes are simpler and more
regular than in many other languages, and stemming algorithms based on heuristics
(rules of thumb) work relatively well \cite{lovins68,porter80,minnen01}.
In an {\em agglutinative} language (e.g., Inuktitut),  many concepts are
combined into a single word, using various prefixes, infixes, and suffixes,
and morphological analysis is complicated. A single word in an agglutinative
language may correspond to a sentence of half a dozen words in English
\cite{johnson03}.

The performance of an information retrieval system is often measured
by precision and recall \cite{manning08}. The {\em precision} of a system is an
estimate of the conditional probability that a document is truly relevant to a query,
if the system says it is relevant. The {\em recall} of a system is
an estimate of the conditional probability that the system will say
that a document is relevant to a query, if it truly is relevant.

In general, normalization increases recall and reduces precision
\cite{kraaij96}. This is natural, given the nature of normalization.
When we remove superficial variations that we believe are irrelevant
to meaning, we make it easier to recognize similarities; we find more
similar things, and so recall increases. But sometimes these superficial
variations have semantic significance; ignoring the variations causes
errors, and so precision decreases. Normalization can also have a positive
effect on precision in cases where variant tokens are infrequent and smoothing
the variations gives more reliable statistics.

If we have a small corpus, we may not be able to afford to be overly selective,
and it may be best to aggressively normalize the text, to increase recall. If we
have a very large corpus, precision may be more important, and we might not want
any normalization. \citeauthor{hull96}~\citeyear{hull96} gives a good analysis of
normalization for information retrieval.

\subsection{Annotation}
\label{subsec:annotation}

Annotation is the inverse of normalization. Just as different strings of
characters may have the same meaning, it also happens that identical strings
of characters may have different meanings, depending on the context.
Common forms of annotation include part-of-speech tagging (marking words
according to their parts of speech), word sense tagging (marking ambiguous
words according to their intended meanings), and parsing (analyzing the grammatical
structure of sentences and marking the words in the sentences according to their
grammatical roles) \cite{manning99}.

Since annotation is the inverse of normalization, we expect it to
decrease recall and increase precision. For example, by tagging {\em program}
as a noun or a verb, we may be able to selectively search for documents that
are about the act of computer programming (verb) instead of documents that discuss
particular computer programs (noun); hence we can increase precision.
However, a document about computer programs (noun) may have something useful
to say about the act of computer programming (verb), even if the document
never uses the verb form of {\em program}; hence we may decrease recall.

Large gains in IR performance have recently been reported as a result
of query annotation with syntactic and semantic information. Syntactic annotation
includes query segmentation \cite{tan08} and part of speech tagging \cite{barr08}.
Examples of semantic annotation are disambiguating abbreviations in queries
\cite{wei08} and finding query keyword associations \cite{lavrenko01,cao05}.

Annotation is also useful for measuring the semantic
similarity of words and concepts (word--context matrices).
For example, \citeauthor{pantel02a}~\citeyear{pantel02a} presented an algorithm
that can discover word senses by clustering row vectors in a
word--context matrix, using contextual information derived from
parsing.

\section{Mathematical Processing for Vector Space Models}
\label{sec:mathematical}

After the text has been tokenized and (optionally) normalized and
annotated, the first step is to generate a matrix of frequencies. Second,
we may want to adjust the weights of the elements in the matrix, because
common words will have high frequencies, yet they are less informative
than rare words. Third, we may want to smooth the matrix, to reduce
the amount of random noise and to fill in some of the zero elements
in a sparse matrix. Fourth, there are many different ways to measure
the similarity of two vectors.

\citeauthor{lowe01}~\citeyear{lowe01} gives a good summary of mathematical processing for
word--context VSMs. He decomposes VSM construction into a similar four-step process:
calculate the frequencies, transform the raw frequency counts, smooth the space
(dimensionality reduction), then calculate the similarities.

\subsection{Building the Frequency Matrix}

An element in a frequency matrix corresponds to an {\em event}: a certain
item (term, word, word pair) occurred in a certain situation (document, context,
pattern) a certain number of times (frequency). At an abstract level, building
a frequency matrix is a simple matter of counting events. In practice,
it can be complicated when the corpus is large.

A typical approach to building a frequency matrix involves two steps. First,
scan sequentially through the corpus, recording events and their frequencies
in a hash table, a database, or a search engine index. Second, use the resulting
data structure to generate the frequency matrix, with a sparse matrix
representation \cite{gilbert92}.

\subsection{Weighting the Elements}
\label{subsec:weighting-elts}

The idea of weighting is to give more weight to surprising events and less
weight to expected events. The hypothesis is that surprising events, if shared
by two vectors, are more discriminative of the similarity between the vectors
than less surprising events. For example, in measuring the semantic similarity
between the words {\em mouse} and {\em rat}, the contexts {\em dissect} and
{\em exterminate} are more discriminative of their similarity than the contexts
{\em have} and {\em like}. In information theory, a surprising event has higher
information content than an expected event \cite{shannon48}. The most popular way
to formalize this idea for term--document matrices is the {\em tf-idf} (term
frequency $\times$ inverse document frequency) family of weighting functions
\cite{sparckjones72}. An element gets a high weight when the corresponding term
is frequent in the corresponding document (i.e., tf is high), but the term is
rare in other documents in the corpus (i.e., df is low, and thus idf
is high). \citeauthor{salton88}~\citeyear{salton88} defined a large family
of tf-idf weighting functions and evaluated them on information retrieval tasks,
demonstrating that tf-idf weighting can yield significant improvements over
raw frequency.

Another kind of weighting, often combined with tf-idf weighting, is
length normalization \cite{singhal96}. In information retrieval, if
document length is ignored, search engines tend to have a bias in favour
of longer documents. Length normalization corrects for this bias.

Term weighting may also be used to correct for correlated terms.
For example, the terms {\em hostage} and {\em hostages} tend to be
correlated, yet we may not want to normalize them to the same term
(as in Section~\ref{subsec:normalization}), because they have
slightly different meanings. As an alternative to normalizing them,
we may reduce their weights when they co-occur in a document
\cite{church95}.

Feature selection may be viewed as a form of weighting, in which
some terms get a weight of zero and hence can be removed from the
matrix. \citeauthor{forman03}~\citeyear{forman03} provides a good study of feature
selection methods for text classification.

An alternative to tf-idf is Pointwise Mutual Information (PMI)
\cite{church89,turney01}, which works well for both word--context matrices
\cite{pantel02a} and term--document matrices \cite{pantel02b}.
A variation of PMI is Positive PMI (PPMI), in which all PMI values
that are less than zero are replaced with zero \cite{niwa94}.
\citeauthor{bullinaria07}~\citeyear{bullinaria07} demonstrated that PPMI
performs better than a wide variety of other weighting approaches
when measuring semantic similarity with word--context matrices.
\citeauthor{turney08b}~\citeyear{turney08b} applied PPMI to pair--pattern matrices.
We will give the formal definition of PPMI here, as an example of an
effective weighting function.

Let $\mathbf{F}$ be a word--context frequency matrix with $n_r$ rows and
$n_c$ columns. The $i$-th row in $\mathbf{F}$ is the row vector
$\mathbf{f}_{i:}$ and the $j$-th column
in $\mathbf{F}$ is the column vector $\mathbf{f}_{:j}$. The row
$\mathbf{f}_{i:}$ corresponds to a word $w_i$ and the column
$\mathbf{f}_{:j}$ corresponds to a context $c_j$. The value of the
element $f_{ij}$ is the number of times that $w_i$ occurs in the
context $c_j$. Let $\mathbf{X}$ be the matrix that results when
PPMI is applied to $\mathbf{F}$. The new matrix $\mathbf{X}$ has
the same number of rows and columns as the raw frequency matrix
$\mathbf{F}$. The value of an element $x_{ij}$ in $\mathbf{X}$
is defined as follows:

{\allowdisplaybreaks 
\begin{align}
p_{ij} & = \frac{f_{ij}}{\sum_{i=1}^{n_r} \sum_{j=1}^{n_c} f_{ij}} \\
p_{i*} & = \frac{\sum_{j=1}^{n_c} f_{ij}}{\sum_{i=1}^{n_r} \sum_{j=1}^{n_c} f_{ij}} \\
p_{*j} & = \frac{\sum_{i=1}^{n_r} f_{ij}}{\sum_{i=1}^{n_r} \sum_{j=1}^{n_c} f_{ij}} \\
\label{eqn:pmi}
{\rm pmi}_{ij} & = \log \left ( \frac{p_{ij}}{p_{i*} p_{*j}} \right ) \\
x_{ij} & =
\left\{
\begin{array}{rl}
{\rm pmi}_{ij} & \mbox{if ${\rm pmi}_{ij} > 0$} \\
0 & \mbox{otherwise}
\end{array}
\right.
\end{align}
} 

In this definition, $p_{ij}$ is the estimated probability that
the word $w_i$ occurs in the context $c_j$, $p_{i*}$ is the estimated
probability of the word $w_i$, and $p_{*j}$ is the estimated probability
of the context $c_j$. If $w_i$ and $c_j$ are statistically independent,
then $p_{i*} p_{*j} = p_{ij}$ (by the definition of independence), and
thus ${\rm pmi}_{ij}$ is zero (since $\log(1) = 0$). The product
$p_{i*} p_{*j}$ is what we would expect for $p_{ij}$ if $w_i$ occurs
in $c_j$ by pure random chance. On the other hand, if there is an
interesting semantic relation between $w_i$ and $c_j$, then
we should expect $p_{ij}$ to be larger than it would be if
$w_i$ and $c_j$ were indepedent; hence we should find that
$p_{ij} > p_{i*} p_{*j}$, and thus ${\rm pmi}_{ij}$ is positive.
This follows from the distributional hypothesis
(see Section~\ref{sec:vsms}). If the word $w_i$ is unrelated
to the context $c_j$, we may find that ${\rm pmi}_{ij}$
is negative. PPMI is designed to give a high value to
$x_{ij}$ when there is an interesting semantic relation between
$w_i$ and $c_j$; otherwise, $x_{ij}$ should have a value of zero,
indicating that the occurrence of $w_i$ in $c_j$ is
uninformative.

A well-known problem of PMI is that it is biased towards infrequent events.
Consider the case where $w_i$ and $c_j$ are statistically dependent
(i.e., they have maximum association). Then $p_{ij} = p_{i*} = p_{*j}$.
Hence (\ref{eqn:pmi}) becomes $\log \left ( 1 / p_{i*} \right )$ and PMI increases
as the probability of word $w_i$ decreases. Several discounting factors have been
proposed to alleviate this problem. An example follows \cite{pantel02a}:

\begin{align}
\delta_{ij} & =
\frac{f_{ij}}{f_{ij}+1} \cdot \frac{\min \left ( \sum_{k=1}^{n_r} f_{kj} ,
\sum_{k=1}^{n_c} f_{ik} \right )}{\min \left ( \sum_{k=1}^{n_r} f_{kj} ,
\sum_{k=1}^{n_c} f_{ik} \right ) + 1} \\
{\rm newpmi}_{ij} & = \delta_{ij} \cdot {\rm pmi}_{ij}
\end{align}

Another way to deal with infrequent events is Laplace smoothing of
the probability estimates, $p_{ij}$, $p_{i*}$, and $p_{*j}$ \cite{turney03a}.
A constant positive value is added to the raw frequencies before calculating
the probabilities; each $f_{ij}$ is replaced with $f_{ij} + k$, for some
$k > 0$. The larger the constant, the greater the smoothing effect.
Laplace smoothing pushes the ${\rm pmi}_{ij}$ values towards zero.
The magnitude of the push (the difference between ${\rm pmi}_{ij}$ with
and without Laplace smoothing) depends on the raw frequency $f_{ij}$.
If the frequency is large, the push is small; if the frequency is small,
the push is large. Thus Laplace smoothing reduces the bias of PMI
towards infrequent events.

\subsection{Smoothing the Matrix}
\label{subsec:smoothing}

The simplest way to improve information retrieval performance is to limit the
number of vector components. Keeping only components representing the most
frequently occurring content words is such a way; however, common words, such
as {\em the} and {\em have}, carry little semantic discrimination power. Simple
component smoothing heuristics, based on the properties of the weighting schemes
presented in Section~\ref{subsec:weighting-elts}, have been shown to both maintain
semantic discrimination power and improve the performance of similarity computations.

Computing the similarity between all pairs of vectors, described in
Section~\ref{subsec:comparing}, is a computationally intensive task. However,
only vectors that share a non-zero coordinate must be compared (i.e., two vectors
that do not share a coordinate are dissimilar). Very frequent context words, such as
the word {\em the}, unfortunately result in most vectors matching a non-zero coordinate.
Such words are precisely the contexts that have little semantic discrimination
power. Consider the pointwise mutual information weighting described in
Section~\ref{subsec:weighting-elts}. Highly weighted dimensions co-occur frequently with
only very few words and are by definition highly discriminating contexts (i.e., they have
very high association with the words with which they co-occur). By keeping only the
context-word dimensions with a PMI above a conservative threshold and setting the others to zero,
\citeauthor{lin98b}~\citeyear{lin98b} showed that the number of comparisons needed to compare vectors greatly
decreases while losing little precision in the similarity score between the top-200 most
similar words of every word. While smoothing the matrix, one computes a reverse index on the
non-zero coordinates. Then, to compare the similarity between a word's context vector and all
other words' context vectors, only those vectors found to match a non-zero component in the
reverse index must be compared. Section~\ref{subsec:efficiency} proposes further optimizations
along these lines.

\citeauthor{deerwester90}~\citeyear{deerwester90} found an elegant way
to improve similarity measurements with a mathematical operation
on the term--document matrix, $\mathbf{X}$, based on linear algebra.
The operation is truncated Singular Value Decomposition (SVD), also
called thin SVD. \citeauthor{deerwester90} briefly mentioned that
truncated SVD can be applied to both document similarity and word similarity,
but their focus was document similarity. \citeauthor{landauer97}~\citeyear{landauer97}
applied truncated SVD to word similarity, achieving human-level scores on
multiple-choice synonym questions from the Test of English as a Foreign Language (TOEFL).
Truncated SVD applied to document similarity is called Latent Semantic
Indexing (LSI), but it is called Latent Semantic Analysis (LSA) when applied
to word similarity.

There are several ways of thinking about how truncated SVD works.
We will first present the math behind truncated SVD and then describe
four ways of looking at it: latent meaning, noise reduction,
high-order co-occurrence, and sparsity reduction.

SVD decomposes $\mathbf{X}$ into the product of three matrices
$\mathbf{U} \mathbf{\Sigma} \mathbf{V}^\mathsf{T}$,
where $\mathbf{U}$ and $\mathbf{V}$ are in column orthonormal
form (i.e., the columns are orthogonal and have unit length,
$\mathbf{U}^\mathsf{T} \mathbf{U} = \mathbf{V}^\mathsf{T} \mathbf{V} = \mathbf{I}$)
and $\mathbf{\Sigma}$ is a diagonal matrix of singular values \cite{golub96}.
If $\mathbf{X}$ is of rank $r$, then $\mathbf{\Sigma}$ is also of rank $r$.
Let ${\mathbf{\Sigma}}_k$, where $k < r$, be the diagonal matrix formed from the top $k$
singular values, and let $\mathbf{U}_k$ and $\mathbf{V}_k$ be the matrices produced
by selecting the corresponding columns from $\mathbf{U}$ and $\mathbf{V}$. The matrix
$\mathbf{U}_k \mathbf{\Sigma}_k \mathbf{V}_k^\mathsf{T}$ is the matrix of rank $k$
that best approximates the original matrix $\mathbf{X}$, in the sense that it
minimizes the approximation errors. That is,
${\bf \hat X} = \mathbf{U}_k \mathbf{\Sigma}_k \mathbf{V}_k^\mathsf{T}$
minimizes $\| {{\bf \hat X} - \mathbf{X}} \|_F$
over all matrices ${\bf \hat X}$ of rank $k$, where $\| \ldots \|_F$
denotes the Frobenius norm \cite{golub96}.

{\bf Latent meaning:} \citeauthor{deerwester90}~\citeyear{deerwester90} and
\citeauthor{landauer97}~\citeyear{landauer97} describe truncated SVD as a method
for discovering latent meaning. Suppose we have a word--context matrix
$\mathbf{X}$. The truncated SVD,
${\bf \hat X} = \mathbf{U}_k \mathbf{\Sigma}_k \mathbf{V}_k^\mathsf{T}$,
creates a low-dimensional linear mapping between row space (words) and
column space (contexts). This low-dimensional mapping captures the latent
(hidden) meaning in the words and the contexts. Limiting the number of latent
dimensions ($k < r$) forces a greater correspondence between words and contexts.
This forced correspondence between words and contexts improves the similarity
measurement.

{\bf Noise reduction:} \citeauthor{rapp03}~\citeyear{rapp03} describes truncated SVD as a
noise reduction technique. We may think of the matrix
${\bf \hat X} = \mathbf{U}_k \mathbf{\Sigma}_k \mathbf{V}_k^\mathsf{T}$
as a smoothed version of the original matrix $\mathbf{X}$. The
matrix $\mathbf{U}_k$ maps the row space (the space spanned by the
rows of $\mathbf{X}$) into a smaller $k$-dimensional space and the
matrix $\mathbf{V}_k$ maps the column space (the space spanned
by the columns of $\mathbf{X}$) into the same $k$-dimensional space.
The diagonal matrix $\mathbf{\Sigma}_k$ specifies the weights in this
reduced $k$-dimensional space. The singular values in $\mathbf{\Sigma}$
are ranked in descending order of the amount of variation in $\mathbf{X}$
that they fit. If we think of the matrix $\mathbf{X}$ as being composed
of a mixture of signal and noise, with more signal than noise, then
$\mathbf{U}_k \mathbf{\Sigma}_k \mathbf{V}_k^\mathsf{T}$ mostly captures
the variation in $\mathbf{X}$ that is due to the signal, whereas the
remaining vectors in $\mathbf{U} \mathbf{\Sigma} \mathbf{V}^\mathsf{T}$
are mostly fitting the variation in $\mathbf{X}$ that is due to the noise.

{\bf High-order co-occurrence:} \citeauthor{landauer97}~\citeyear{landauer97}
also describe truncated SVD as a method for discovering high-order
co-occurrence. Direct co-occurrence (first-order co-occurrence) is
when two words appear in identical contexts. Indirect co-occurrence
(high-order co-occurrence) is when two words appear in {\em similar}
contexts. Similarity of contexts may be defined recursively in terms of
lower-order co-occurrence. \citeauthor{lemaire06}~\citeyear{lemaire06}
demonstrate that truncated SVD can discover high-order co-occurrence.

{\bf Sparsity reduction:} In general, the matrix $\mathbf{X}$ is
very sparse (mostly zeroes), but the truncated SVD,
${\bf \hat X} = \mathbf{U}_k \mathbf{\Sigma}_k \mathbf{V}_k^\mathsf{T}$,
is dense. Sparsity may be viewed as a problem of insufficient data:
with more text, the matrix $\mathbf{X}$ would have fewer zeroes,
and the VSM would perform better on the chosen task. From this perspective,
truncated SVD is a way of simulating the missing text, compensating
for the lack of data \cite{vozalis03}.

These different ways of viewing truncated SVD are compatible with
each other; it is possible for all of these perspectives to be correct.
Future work is likely to provide more views of SVD and perhaps a
unifying view.

A good C implementation of SVD for large sparse matrices is Rohde's
SVDLIBC.\footnote{SVDLIBC is available at
http://tedlab.mit.edu/$\scriptstyle\sim$dr/svdlibc/.} Another approach
is Brand's~\citeyear{brand06} incremental truncated SVD algorithm.\footnote{MATLAB
source code is available at
http://web.mit.edu/$\scriptstyle\sim$wingated/www/resources.html.}
Yet another approach is Gorrell's~\citeyear{gorrell06} Hebbian algorithm for
incremental truncated SVD. Brand's and Gorrell's algorithms both introduce
interesting new ways of handling missing values, instead of treating them
as zero values.

For higher-order tensors, there are operations that are analogous
to truncated SVD, such as parallel factor analysis (PARAFAC) \cite{harshman70},
canonical decomposition (CANDECOMP) \cite{carroll70} (equivalent to PARAFAC but
discovered independently), and Tucker decomposition \cite{tucker66}.
For an overview of tensor decompositions, see the surveys of
\citeauthor{kolda09}~\citeyear{kolda09} or \citeauthor{acar09}~\citeyear{acar09}.
\citeauthor{turney07}~\citeyear{turney07} gives an empirical evaluation of how well four
different Tucker decomposition algorithms scale up for large sparse third-order
tensors. A low-RAM algorithm, Multislice Projection, for large sparse
tensors is presented and evaluated.\footnote{MATLAB source code is available at
http://www.apperceptual.com/multislice/.}

Since the work of \citeauthor{deerwester90}~\citeyear{deerwester90},
subsequent research has discovered many alternative matrix smoothing processes,
such as Nonnegative Matrix Factorization (NMF) \cite{lee99}, Probabilistic
Latent Semantic Indexing (PLSI) \cite{hofmann99}, Iterative Scaling (IS)
\cite{ando00}, Kernel Principal Components Analysis (KPCA) \cite{scholkopf97},
Latent Dirichlet Allocation (LDA) \cite{blei03}, and Discrete Component Analysis
(DCA) \cite{buntine06}.

The four perspectives on truncated SVD, presented above, apply equally
well to all of these more recent matrix smoothing algorithms.
These newer smoothing algorithms tend to be more computationally intensive
than truncated SVD, but they attempt to model word frequencies better
than SVD. Truncated SVD implicitly assumes that the elements in $\mathbf{X}$
have a Gaussian distribution. Minimizing the the Frobenius norm
$\| {{\bf \hat X} - \mathbf{X}} \|_F$ will minimize the noise, if the noise
has a Gaussian distribution. However, it is known that word frequencies do not
have Gaussian distributions. More recent algorithms are based on more realistic
models of the distribution for word frequencies.\footnote{In our experience,
${\rm pmi}_{ij}$ appears to be approximately Gaussian, which may explain why
PMI works well with truncated SVD, but then PPMI is puzzling, because it is
less Gaussian than PMI, yet it apparently yields better semantic models than PMI.}

\subsection{Comparing the Vectors}
\label{subsec:comparing}

The most popular way to measure the similarity of two frequency vectors
(raw or weighted) is to take their cosine. Let $\mathbf{x}$ and $\mathbf{y}$
be two vectors, each with $n$ elements.

\begin{align}
\mathbf{x} & = \left \langle x_1, x_2, \ldots, x_n \right \rangle \\
\mathbf{y} & = \left \langle y_1, y_2, \ldots, y_n \right \rangle
\end{align}

\noindent The cosine of the angle $\theta$ between $\mathbf{x}$ and $\mathbf{y}$
can be calculated as follows:

\begin{align}
\label{eqn:cosine}
{\rm cos}(\mathbf{x},\mathbf{y})
& = \frac{{\sum_{i = 1}^n {x_i \cdot y_i} }}
{{\sqrt {\sum_{i = 1}^n {x_i^2 \cdot \sum_{i = 1}^n y_i^2  } } }} \\
& = \frac{{\mathbf{x}  \cdot \mathbf{y} }}
{{\sqrt {\mathbf{x} \cdot \mathbf{x} } \cdot
\sqrt {\mathbf{y}  \cdot \mathbf{y} } }} \\
& = \frac{{\mathbf{x}}}{{\left\| {\mathbf{x} } \right\| }}
\cdot \frac{{\mathbf{y}}}{{\left\| {\mathbf{y} } \right\| }}
\end{align}

\noindent In other words, the cosine of the angle between two vectors is
the inner product of the vectors, after they have been normalized to unit length.
If $\mathbf{x}$ and $\mathbf{y}$ are frequency vectors for
words, a frequent word will have a long vector and a rare word will have
a short vector, yet the words might be synonyms. Cosine captures the idea
that the length of the vectors is irrelevant; the important thing is the
angle between the vectors.

The cosine ranges from $-1$ when the vectors point in opposite directions
($\theta$ is 180 degrees) to $+1$ when they point in the same direction
($\theta$ is 0 degrees). When the vectors are orthogonal ($\theta$ is 90 degrees),
the cosine is zero. With raw frequency vectors, which necessarily cannot
have negative elements, the cosine cannot be negative, but weighting and smoothing
often introduce negative elements. PPMI weighting does not yield negative
elements, but truncated SVD can generate negative elements, even when the
input matrix has no negative values.

A measure of distance between vectors can easily be converted to a measure
of similarity by inversion (\ref{eqn:inversion}) or subtraction
(\ref{eqn:subtraction}).

\begin{align}
\label{eqn:inversion}
{\rm sim}(\mathbf{x},\mathbf{y}) & = 1 / {\rm dist}(\mathbf{x},\mathbf{y}) \\
\label{eqn:subtraction}
{\rm sim}(\mathbf{x},\mathbf{y}) & = 1 - {\rm dist}(\mathbf{x},\mathbf{y})
\end{align}

\noindent Many similarity measures have been proposed in both IR \cite{jones87}
and lexical semantics circles \cite{lin98b,dagan99,lillianlee99,weeds04}. It is
commonly said in IR that, properly normalized, the difference in retrieval performance
using different measures is insignificant \cite{vanrijsbergen79}. Often the vectors
are normalized in some way (e.g., unit length or unit probability) before applying
any similarity measure.

Popular geometric measures of vector distance
include Euclidean distance and Manhattan distance. Distance measures
from information theory include Hellinger, Bhattacharya, and Kullback-Leibler.
\citeauthor{bullinaria07}~\citeyear{bullinaria07} compared these five distance
measures and the cosine similarity measure on four different tasks involving
word similarity. Overall, the best measure was cosine. Other popular measures
are the Dice and Jaccard coefficients \cite{manning08}.

\citeauthor{lillianlee99}~\citeyear{lillianlee99} proposed that, for finding word similarities,
measures that focused more on overlapping coordinates and less on the importance of negative
features (i.e., coordinates where one word has a nonzero value and the other has a
zero value) appear to perform better. In Lee's experiments, the Jaccard, Jensen-Shannon,
and L1 measures seemed to perform best. \citeauthor{weeds04}~\citeyear{weeds04} studied the
linguistic and statistical properties of the similar words returned by various
similarity measures and found that the measures can be grouped into three classes:

\begin{myenumerate}

\item high-frequency sensitive measures (cosine, Jensen-Shannon, $\alpha$-skew, recall),

\item low-frequency sensitive measures (precision), and

\item similar-frequency sensitive methods (Jaccard, Jaccard+MI, Lin, harmonic mean).

\end{myenumerate}

\noindent Given a word $w_0$, if we use a high-frequency sensitive measure
to score other words $w_i$ according to their similarity with $w_0$, higher frequency
words will tend to get higher scores than lower frequency words. If we use
a low-frequency sensitive measure, there will be a bias towards lower frequency
words. Similar-frequency sensitive methods prefer a word $w_i$ that has
approximately the same frequency as $w_0$. In one experiment on determining the
compositionality of collocations, high-frequency sensitive measures outperformed
the other classes \cite{weeds04}. We believe that determining the most appropriate
similarity measure is inherently dependent on the similarity task, the sparsity of
the statistics, the frequency distribution of the elements being compared, and the
smoothing method applied to the matrix.

\subsection{Efficient Comparisons}
\label{subsec:efficiency}

Computing the similarity between all rows (or columns) in a large matrix is a
non-trivial problem, with a worst case cubic running time $O(n_r^2n_c)$, where $n_r$
is the number of rows and $n_c$ is the number of columns (i.e., the dimensionality of
the feature space). Optimizations and parallelization are often necessary.

\subsubsection{Sparse-Matrix Multiplication}
\label{subsec:sparse-matrix-mult}

One optimization strategy is a generalized sparse-matrix multiplication
approach \cite{sarawagi04}, which is based on the observation that a scalar product
of two vectors depends only on the coordinates for which both vectors have nonzero
values. Further, we observe that most commonly used similarity measures for vectors
$\mathbf{x}$ and $\mathbf{y}$, such as cosine, overlap, and Dice,
can be decomposed into three
values: one depending only on the nonzero values of $\mathbf{x}$, another depending
only on the nonzero values of $\mathbf{y}$, and the third depending on the nonzero
coordinates shared both by $\mathbf{x}$ and $\mathbf{y}$. More formally, commonly used
similarity scores, ${\rm sim}(\mathbf{x},\mathbf{y})$, can be
expressed as follows:

\begin{equation}
{\rm sim}(\mathbf{x},\mathbf{y}) = f_0 \left (
\textstyle{\sum_{i=1}^{n}} f_1(x_i,y_i),
f_2(\mathbf{x}), f_3(\mathbf{y}) \right )
\label{eqn:sim}
\end{equation}

\noindent For example, the cosine measure, ${\rm cos}(\mathbf{x},\mathbf{y})$,
defined in (\ref{eqn:cosine}), can be expressed in this model as follows:

\begin{align}
{\rm cos}(\mathbf{x},\mathbf{y}) & = f_0 \left (
\textstyle{\sum_{i=1}^{n}} f_1(x_i,y_i),
f_2(\mathbf{x}), f_3(\mathbf{y}) \right ) \\
f_0(a,b,c) & = \frac{a}{b \cdot c} \\
f_1(a,b) & = a \cdot b \\
f_2(\mathbf{a}) & = f_3(\mathbf{a}) = \sqrt{\textstyle{\sum_{i=1}^{n}} a_i^2}
\end{align}

Let $\mathbf{X}$ be a matrix for which we want to compute the pairwise similarity,
${\rm sim}(\mathbf{x},\mathbf{y})$, between
all rows or all columns $\mathbf{x}$ and $\mathbf{y}$.
Efficient computation of the similarity matrix $\mathbf{S}$ can be achieved by
leveraging the fact that ${\rm sim}(\mathbf{x},\mathbf{y})$
is determined solely by the nonzero
coordinates shared by $\mathbf{x}$ and $\mathbf{y}$
(i.e., $f_1(0,x_i)$ = $f_1(x_i,0)$ = 0 for any $x_i$) and that most of the vectors are
very sparse. In this case, calculating $f_1(x_i,y_i)$ is only required when both
vectors have a shared nonzero coordinate, significantly reducing the cost of
computation. Determining which vectors share a nonzero coodinate can easily be
achieved by first building an inverted index for the coordinates. During indexing,
we can also precompute $f_2(\mathbf{x})$ and $f_3(\mathbf{y})$ without changing the
algorithm complexity. Then, for each vector $\mathbf{x}$ we retrieve in constant time,
from the index, each vector $\mathbf{y}$ that shares a nonzero coordinate
with $\mathbf{x}$ and we apply $f_1(x_i,y_i)$ on the shared coordinates $i$.
The computational cost of this algorithm is $\sum_{i} N_i^2$ where $N_i$ is the
number of vectors that have a nonzero $i$-th coordinate. Its worst case time
complexity is $O(ncv)$ where $n$ is the number of vectors to be compared, $c$ is
the maximum number of nonzero coordinates of any vector, and $v$ is the number of
vectors that have a nonzero $i$-th coordinate where $i$ is the coordinate which
is nonzero for the most vectors. In other words, the algorithm is efficient only
when the density of the coordinates is low. In our own experiments of computing the
semantic similarity between all pairs of words in a large web crawl, we observed
near linear average running time complexity in $n$.

The computational cost can be reduced further by leveraging the element weighting
techniques described in Section~\ref{subsec:weighting-elts}. By setting to zero
all coordinates that have a low PPMI, PMI or tf-idf score, the coordinate density
is dramatically reduced at the cost of losing little discriminative power. In this
vein, \citeauthor{bayardo07}~\citeyear{bayardo07} described a strategy that omits the
coordinates with the highest number of nonzero values. Their algorithm gives a
significant advantage only when we are interested in finding solely the similarity
between highly similar vectors.

\subsubsection{Distributed Implementation using MapReduce}
\label{subsec:mapreduce}

The algorithm described in Section~\ref{subsec:sparse-matrix-mult} assumes that the
matrix $\mathbf{X}$ can fit into memory, which for large $\mathbf{X}$ may be
impossible. Also, as each element of $\mathbf{X}$ is processed independently, running
parallel processes for non-intersecting subsets of $\mathbf{X}$ makes the processing
faster. \citeauthor{elsayed08}~\citeyear{elsayed08} proposed a MapReduce implementation
deployed using Hadoop, an open-source software package implementing the MapReduce
framework and distributed file system.\footnote{Hadoop is available for download
at http://lucene.apache.org/hadoop/.} Hadoop has been shown to scale
to several thousands of machines, allowing users to write simple
code, and to seamlessly manage the sophisticated parallel execution of the code.
\citeauthor{dean08}~\citeyear{dean08} provide a good primer on MapReduce programming.

The MapReduce model's Map step is used to start $m \times n$ Map tasks in parallel,
each caching one $m$-th part of $\mathbf{X}$ as an inverted index and
streaming one $n$-th part of $\mathbf{X}$ through it. The actual inputs
are read by the tasks directly from HDFS (Hadoop Distributed File System). The value of $m$ is
determined by the amount of memory dedicated for the inverted index, and $n$ should
be determined by trading off the fact that, as $n$ increases, more parallelism can be
obtained at the increased cost of building the same inverted index $n$ times.

The similarity algorithm from Section~\ref{subsec:sparse-matrix-mult} runs in each
task of the Map step of a MapReduce job. The Reduce step
groups the output by the rows (or columns) of $\mathbf{X}$.

\subsubsection{Randomized Algorithms}
\label{subsec:randomized-algorithms}

Other optimization strategies use randomized techniques to approximate various
similarity measures. The aim of randomized algorithms is to improve computational
efficiency (memory and time) by projecting high-dimensional vectors into a
low-dimensional subspace. Truncated SVD performs such a projection, but SVD can
be computationally intensive.\footnote{However, there are efficient forms of SVD
\cite{brand06,gorrell06}.} The insight of randomized techniques is that
high-dimensional vectors can be randomly projected into a low-dimensional subspace
with relatively little impact on the final similarity scores.
Significant reductions in computational cost have been reported
with little average error to computing the true similarity scores, especially in
applications such as word similarity where we are interested in only the top-$k$ most
similar vectors to each vector \cite{ravichandran05,gorman06}.

{\em Random Indexing}, an approximation technique based on Sparse Distributed Memory
\cite{kanerva93}, computes the pairwise similarity between all rows (or vectors) of a
matrix with complexity $O(n_rn_c\delta_1)$, where $\delta_1$ is a fixed constant
representing the length of the index vectors assigned to each column.
The value of $\delta_1$ controls the tradeoff of accuracy
versus efficiency. The elements of each index vector are
mostly zeros with a small number of randomly assigned $+1$'s and $-1$'s. The cosine measure
between two rows $\mathbf{r}_1$ and $\mathbf{r}_2$ is then approximated by computing
the cosine between two {\em fingerprint} vectors, $\mathbf{fingerprint}(\mathbf{r}_1)$ and
$\mathbf{fingerprint}(\mathbf{r}_2)$, where $\mathbf{fingerprint(r)}$ is computed by summing
the index vectors of each non-unique coordinate of $\mathbf{r}$. {\em Random Indexing} was
shown to perform as well as LSA on a word synonym selection task \cite{karlgren01}.

Locality sensitive hashing (LSH) \cite{broder97} is another technique that
approximates the similarity matrix with complexity $O(n_r^2\delta_2)$, where
$\delta_2$ is a constant number of random projections, which controls the accuracy
versus efficiency tradeoff.\footnote{LSH stems from work by \citeauthor{rabin81}
\citeyear{rabin81}, who proposed the use of hash functions from random irreducible
polynomials to create short fingerprints of collections of documents. Such
techniques are useful for many tasks, such as removing duplicate
documents ({\em deduping}) in a web crawl.}
LSH is a general class of techniques for defining functions that map vectors (rows
or columns) into short signatures or fingerprints, such that two similar vectors are
likely to have similar fingerprints. Definitions of LSH functions include the Min-wise
independent function, which preserves the Jaccard similarity between vectors
\cite{broder97}, and functions that preserve the cosine similarity between vectors
\cite{charikar02}. On a word similarity task, \citeauthor{ravichandran05}
\citeyear{ravichandran05} showed that, on average, over 80\% of the top-10 similar
words of random words are found in the top-10 results using Charikar's functions,
and that the average cosine error is 0.016 (using
$\delta_2$ = 10,000 random projections). \citeauthor{gorman06}~\citeyear{gorman06}
provide a detailed comparison of Random Indexing and LSH on a distributional
similarity task. On the BNC corpus, LSH outperformed Random Indexing; however,
on a larger corpora combining BNC, the Reuters Corpus, and most of the English
news holdings of the LDC in 2003, Random Indexing outperformed LSH in both
efficiency and accuracy.

\subsection{Machine Learning}
\label{subsec:ml}

If the intended application for a VSM is clustering or classification,
a similarity measure such as cosine (Section~\ref{subsec:comparing}) may
be used. For classification, a nearest-neighbour algorithm can use
cosine as a measure of nearness \cite{dasarathy91}. For clustering, a similarity-based
clustering algorithm can use cosine as a measure of similarity \cite{jain99}.
However, there are many machine learning algorithms that can work directly
with the vectors in a VSM, without requiring an external similarity measure,
such as cosine. In effect, such machine learning algorithms implicitly use
their own internal approaches to measuring similarity.

Any machine learning algorithm that works with real-valued vectors
can use vectors from a VSM \cite{witten05}. Linguistic processing
(Section~\ref{sec:linguistic}) and mathematical processing
(Section~\ref{sec:mathematical}) may still be necessary, but the
machine learning algorithm can handle vector comparison
(Sections \ref{subsec:comparing} and \ref{subsec:efficiency}).

In addition to unsupervised (clustering) and supervised (classification)
machine learning, vectors from a VSM may also be used in semi-supervised
learning \cite{ando05,collobert08}. In general, there is nothing unique
to VSMs that would compel a choice of one machine learning algorithm
over another, aside from the algorithm's performance on the given task.
Therefore we refer our readers to the machine learning literature
\cite{witten05}, since we have no advice that is specific to VSMs.

\section{Three Open Source VSM Systems}
\label{sec:three-systems}

To illustrate the three types of VSMs discussed in Section~\ref{sec:vsms},
this section presents three open source systems, one for each VSM type.
We have chosen to present open source systems so that interested
readers can obtain the source code to find out more about the systems
and to apply the systems in their own projects. All three systems are
written in Java and are designed for portability and ease of use.

\subsection{The Term--Document Matrix: Lucene}
\label{subsec:lucene}

Lucene\footnote{Apache Lucene is available for download at http://lucene.apache.org/.}
is an open source full-featured text search engine library supported by the Apache
Software Foundation. It is arguably the most ubiquitous implementation of a term--document
matrix, powering many search engines such as at CNET, SourceForge, Wikipedia, Disney, AOL
and Comcast. Lucene offers efficient storage, indexing, as well as retrieval and ranking
functionalities. Although it is primarily used as a term--document matrix, it generalizes
to other VSMs.

Content, such as webpages, PDF documents, images, and video, are programmatically
decomposed into {\em fields} and stored in a database. The database implements the
term--document matrix, where content corresponds to documents and fields correspond to
terms. Fields are stored in the database and indices are computed on the field values.
Lucene uses fields as a generalization of content terms, allowing any other {\em string}
or {\em literal} to index documents. For example, a webpage could be indexed by all the
terms it contains, and also by the anchor texts pointing to it, its host name, and the
semantic classes in which it is classified (e.g., spam, product review, news, etc.). The
webpage can then be retrieved by search terms matching any of these {\em fields}.

Columns in the term--document matrix consist of all the fields of a particular instance
of content (e.g., a webpage). The rows consist of all instances of content in the index.
Various statistics such as {\em frequency} and {\em tf-idf} are stored in the matrix.
The developer defines the fields in a schema and identifies those to be indexed by
Lucene. The developer also optionally defines a content ranking function for each
indexed field.

Once the index is built, Lucene offers functionalities for retrieving content. Users
can issue many query types such as phrase queries, wildcard queries, proximity queries,
range queries (e.g., date range queries), and field-restricted queries. Results can be
sorted by any field and index updates can occur simultaneously during searching. Lucene's
index can be directly loaded into a Tomcat webserver and it offers APIs for common
programming languages. Solr,\footnote{Apache Solr is available for download at http://lucene.apache.org/solr/.} a separate Apache Software Foundation project, is an
open source enterprise webserver for searching a Lucene index and presenting search
results. It is a full-featured webserver providing functionalities such as XML/HTTP
and JSON APIs, hit highlighting, faceted search, caching, and replication.

A simple recipe for creating a web search service, using Nutch, Lucene
and Solr, consists of crawling a set of URLs (using Nutch), creating
a term--document matrix index (using Lucene), and serving search results (using Solr).
Nutch,\footnote{Apache Nutch is available for download at http://lucene.apache.org/nutch/.}
the Apache Software Foundation open source web search software, offers functionality for
crawling the web from a seed set of URLs, for building a link-graph of the web crawl, and
for parsing web documents such as HTML pages. A good set of seed URLs for Nutch can be
downloaded freely from the Open Directory Project.\footnote{See http://www.dmoz.org/.}
Crawled pages are HTML-parsed, and they are then indexed by Lucene. The
resulting indexed collection is then queried and served through a Solr installation
with Tomcat.

For more information on Lucene, we recommend \citeauthor{gospodnetic04}'s
\citeyear{gospodnetic04} book. \citeauthor{konchady08}~\citeyear{konchady08} explains how to
integrate Lucene with LingPipe and GATE for sophisticated semantic
processing.\footnote{Information about LingPipe is available at http://alias-i.com/lingpipe/.
The GATE (General Architecture for Text Engineering) home page is at
http://gate.ac.uk/.}

\subsection{The Word--Context Matrix: Semantic Vectors}

Semantic Vectors\footnote{Semantic Vectors is a software package for measuring
word similarity, available under the Simplified BSD License at
http://code.google.com/p/semanticvectors/.} is an open source project implementing
the random projection approach to measuring word similarity (see
Section~\ref{subsec:randomized-algorithms}). The package uses Lucene
to create a term--document matrix, and it then creates vectors from Lucene's
term--document matrix, using random projection for dimensionality reduction.
The random projection vectors can be used, for example, to measure the semantic
similarity of two words or to find the words that are most similar to a given word.

The idea of random projection is to take high-dimensional vectors and
randomly project them into a relatively low-dimensional space \cite{sahlgren05}.
This can be viewed as a kind of smoothing operation (Section~\ref{subsec:smoothing}),
but the developers of the Semantic Vectors package emphasize the simplicity and
efficiency of random projection (Section~\ref{subsec:efficiency}), rather than
its smoothing ability. They argue that other matrix smoothing algorithms might
smooth better, but none of them perform as well as random indexing, in terms of
the computational complexity of building a smooth matrix and incrementally
updating the matrix when new data arrives \cite{widdows08}. Their aim
is to encourage research and development with semantic vectors by creating
a simple and efficient open source package.

The Semantic Vectors package is designed to be convenient to use, portable,
and easy to extend and modify. The design of the software
incorporates lessons learned from the earlier Stanford Infomap
project.\footnote{See http://infomap-nlp.sourceforge.net/.}
Although the default is to generate random projection vectors, the
system has a modular design that allows other kinds of word--context
matrices to be used instead of random projection matrices.

The package supports two basic functions: building a word--context matrix
and searching through the vectors in the matrix. In addition to generating
word vectors, the building operation can generate document vectors
by calculating weighted sums of the word vectors for the words in
each document. The searching operation can be used to search for similar
words or to search for documents that are similar to a query. A query
can be a single word or several words can be combined, using various
mathematical operations on the corresponding vectors. The mathematical
operations include vector negation and disjunction, based on quantum
logic \cite{widdows04}. \citeauthor{widdows08}~\citeyear{widdows08} provide a
good summary of the Semantic Vectors software.

\subsection{The Pair--Pattern Matrix: Latent Relational Analysis in S-Space}

Latent Relational Analysis\footnote{Latent Relational Analysis is part of the
{\em S-Space} package and is distributed under the GNU General Public License
version 2. It is available at http://code.google.com/p/airhead-research/.
At the time of writing, the LRA module was under development.}
(LRA) is an open source project implementing the pair--pattern matrix. It is a
component of the {\em S-Space} package, a library of tools for building and
comparing different semantic spaces.

LRA takes as input a textual corpus and a set of word pairs. A pair--pattern
matrix is built by deriving lexical patterns that link together the word pairs
in the corpus. For example, consider the word pair $\langle$Korea, Japan$\rangle$
and the following retrieved matching sentences:

\begin{myitemize}

\item {\em Korea looks to new Japan prime minister's effect on Korea-Japan relations.}

\item {\em What channel is the Korea vs. Japan football game?}

\end{myitemize}

\noindent From these two sentences, LRA extracts two patterns: ``$X$ looks to new $Y$''
and ``$X$ vs. $Y$''. These patterns become two columns in the pair--pattern matrix, and
the word pair $\langle$Korea, Japan$\rangle$ becomes a row. Pattern frequencies are
counted and then smoothed using SVD (see Section~\ref{subsec:smoothing}).

In order to mitigate the sparseness of occurrences of word pairs, a thesaurus such as
WordNet is used to expand the seed word pairs to alternatives. For example the pair
$\langle$Korea, Japan$\rangle$ may be expanded to include $\langle$South Korea,
Japan$\rangle$, $\langle$Republic of Korea, Japan$\rangle$, $\langle$Korea,
Nippon$\rangle$, $\langle$South Korea, Nippon$\rangle$, and $\langle$Republic
of Korea, Nippon$\rangle$.

LRA uses Lucene (see Section~\ref{subsec:lucene}) as its backend to store the
matrix, index it, and serve its contents. For a detailed description of the LRA
algorithm, we suggest \citeauthor{turney06}'s \citeyear{turney06} paper.

\section{Applications}
\label{sec:applications}

In this section, we will survey some of the semantic applications for
VSMs. We will aim for breadth, rather than depth; readers who want more
depth should consult the references. Our goal is to give the reader
an impression of the scope and flexibility of VSMs for semantics.
The following applications are grouped according to the type of matrix
involved: term--document, word--context, or pair--pattern. Note that this
section is not exhaustive; there are many more references and applications
than we have space to list here.

\subsection{Term--Document Matrices}
\label{subsec:term--document-apps}

Term--document matrices are most suited to measuring the semantic similarity
of documents and queries (see Section~\ref{subsec:term--document-vsms}). The
usual measure of similarity is the cosine of column vectors in a weighted
term--document matrix. There are a variety of applications for measures of
document similarity.

{\bf Document retrieval:} The term--document matrix
was first developed for document retrieval \cite{salton75},
and there is now a large body of literature on the VSM for document retrieval
\cite{manning08}, including several journals and conferences
devoted to the topic. The core idea is, given a query, rank the
documents in order of decreasing cosine of the angles between the
query vector and the document vectors \cite{salton75}. One variation on the theme
is cross-lingual document retrieval, where a query in one language is used
to retrieve a document in another language \cite{landauer90}. An important
technical advance was the discovery that smoothing the term--document matrix by
truncated SVD can improve precision and recall \cite{deerwester90},
although few commercial systems use smoothing, due to the computational
expense when the document collection is large and dynamic. Random
indexing \cite{sahlgren05} or incremental SVD \cite{brand06} may help
to address these scaling issues. Another important development in
document retrieval has been the addition of collaborative filtering,
in the form of PageRank \cite{brin98}.

{\bf Document clustering:} Given a measure of document similarity, we can
cluster the documents into groups, such that similarity tends to be high within
a group, but low across groups \cite{manning08}. The clusters may be partitional
(flat) \cite{cutting92,pantel02b} or they may have a hierarchical structure (groups of groups)
\cite{zhao02}; they may be non-overlapping (hard) \cite{croft77}
or overlapping (soft) \cite{zamir99}. Clustering algorithms
also differ in how clusters are compared and abstracted. With single-link clustering,
the similarity between two clusters is the maximum of the similarities between
their members. Complete-link clustering uses the minimum of the similarities
and average-link clustering uses the average of the similarities \cite{manning08}.

{\bf Document classification:} Given a training set of documents with
class labels and a testing set of unlabeled documents, the task of document
classification is to learn from the training set how to assign labels to the testing
set \cite{manning08}. The labels may be the topics of the documents \cite{sebastiani02},
the sentiment of the documents (e.g., positive versus negative product
reviews) \cite{pang02,kim06}, spam versus non-spam \cite{sahami98,pantel98}, or any
other labels that might be inferred from the words in the documents. When we classify
documents, we are implying that the documents in a class are similar in some way;
thus document classification implies some notion of document similarity, and most
machine learning approaches to document classification involve a term--document
matrix \cite{sebastiani02}. A measure of document similarity, such as cosine,
can be directly applied to document classification by using a nearest-neighbour
algorithm \cite{yang99}.

{\bf Essay grading:} Student essays may be automatically graded by
comparing them to one or more high-quality reference essays on the
given essay topic \cite{wolfe98,foltz99}. The student essays and
the reference essays can be compared by their cosines in a term--document
matrix. The grade that is assigned to a student essay is proportional
to its similarity to one of the reference essays; a student essay that is highly
similar to a reference essay gets a high grade.

{\bf Document segmentation:} The task of document segmentation is to
partition a document into sections, where each section focuses on a
different subtopic of the document \cite{hearst97,choi00}. We may treat
the document as a series of blocks, where a block is a sentence or
a paragraph. The problem is to detect a topic shift from one block to
the next. \citeauthor{hearst97}~\citeyear{hearst97} and \citeauthor{choi00}~\citeyear{choi00}
both use the cosine between columns in a word--block frequency matrix to measure
the semantic similarity of blocks. A topic shift is signaled by a drop
in the cosine between consecutive blocks. The word--block matrix can
be viewed as a small term--document matrix, where the corpus is a single
document and the documents are blocks.

{\bf Question answering:} Given a simple question, the task in
Question Answering (QA) is to find a short answer to the question by searching
in a large corpus. A typical question is, ``How many calories are there
in a Big Mac?'' Most algorithms for QA have four components, question analysis,
document retrieval, passage retrieval, and answer extraction \cite{tellex03,dang06}.
Vector-based similarity measurements are often used for both document retrieval
and passage retrieval \cite{tellex03}.

{\bf Call routing:} \citeauthor{chucarroll99}~\citeyear{chucarroll99}
present a vector-based system for automatically routing telephone calls,
based on the caller's spoken answer to the question, ``How may I direct your call?''
If the caller's answer is ambiguous, the system automatically generates a
question for the caller, derived from the VSM, that prompts the caller
for further information.

\subsection{Word--Context Matrices}
\label{subsec:word--context-apps}

Word--context matrices are most suited to measuring the semantic similarity of
words (see Section~\ref{subsec:word--context-vsms}). For example, we can measure
the similarity of two words by the cosine of the angle between their corresponding
row vectors in a word--context matrix. There are many applications for measures of
word similarity.

{\bf Word similarity:} \citeauthor{deerwester90}~\citeyear{deerwester90}
discovered that we can measure word similarity by comparing row vectors in a
term--document matrix. \citeauthor{landauer97}~\citeyear{landauer97} evaluated
this approach with 80 multiple-choice synonym
questions from the Test of English as a Foreign Language (TOEFL),
achieving human-level performance (64.4\% correct for the word--context matrix
and 64.5\% for the average non-English US college applicant). The documents used
by \citeauthor{landauer97} had an average length
of 151 words, which seems short for a document, but long for the context
of a word. Other researchers soon switched to much shorter lengths, which
is why we prefer to call these {\em word--context} matrices, instead of
{\em term--document} matrices. \citeauthor{lund96}~\citeyear{lund96} used
a context window of ten words. \citeauthor{schutze98}~\citeyear{schutze98} used a
fifty-word window ($\pm$25 words, centered on the target word).
\citeauthor{rapp03}~\citeyear{rapp03} achieved 92.5\% correct on the 80 TOEFL questions,
using a four-word context window ($\pm$2 words, centered on the target word,
after removing stop words). The TOEFL results suggest that performance
improves as the context window shrinks. It seems that the immediate context of
a word is much more important than the distant context for determining
the meaning of the word.

{\bf Word clustering:} \citeauthor{pereira93}~\citeyear{pereira93}
applied soft hierarchical clustering to row-vectors in a word--context
matrix. In one experiment, the words were nouns and the contexts
were verbs for which the given nouns were direct objects. In another
experiment, the words were verbs and the contexts were nouns that were
direct objects of the given verbs. \citeauthor{schutze98}'s
\citeyear{schutze98} seminal word sense discrimination model
used hard flat clustering for row-vectors in a word--context matrix, where
the context was given by a window of $\pm$25 words, centered on the target word.
\citeauthor{pantel02a}~\citeyear{pantel02a} applied soft flat clustering to a
word--context matrix, where the context was based on parsed text.
These algorithms are able to discover different senses of polysemous
words, generating different clusters for each sense. In effect, the
different clusters correspond to the different concepts that underlie
the words.

{\bf Word classification:}
\citeauthor{turney03a}~\citeyear{turney03a} used a word--context
matrix to classify words as positive ({\em honest}, {\em intrepid})
or negative ({\em disturbing}, {\em superfluous}). They used the General
Inquirer (GI) lexicon \cite{stone66} to evaluate their algorithms.
The GI lexicon includes 11,788 words, labeled with 182 categories related to
opinion, affect, and attitude.\footnote{The GI lexicon is available at
http://www.wjh.harvard.edu/$\scriptstyle\sim$inquirer/spreadsheet\_guide.htm.}
\citeauthor{turney03a}  hypothesize that all 182 categories
can be discriminated with a word--context matrix.

{\bf Automatic thesaurus generation:} WordNet is a popular tool for research
in natural language processing \cite{fellbaum98}, but creating and maintaing
such lexical resources is labour intensive, so it is natural to wonder whether
the process can be automated to some degree.\footnote{WordNet is available at
http://wordnet.princeton.edu/.} This task can seen as an instance
of word clustering (when the thesaurus is generated from scratch) or
classification (when an existing thesaurus is automatically extended), but
it is worthwhile to consider the task of automatic thesaurus generation
separately from clustering and classification, due to the specific
requirements of a thesaurus, such as the particular kind of similarity
that is appropriate for a thesaurus (see Section~\ref{subsec:similarities}).
Several researchers have used word--context matrices specifically for the
task of assisting or automating thesaurus generation
\cite{crouch88,grefenstette94,ruge97,pantel02a,curran02}.

{\bf Word sense disambiguation:} A typical Word Sense Disambiguation (WSD)
system \cite{aggire06,pedersen06} uses a feature vector representation in
which each vector corresponds to a token of a word, not a type (see
Section~\ref{subsec:type-token}). However, \citeauthor{leacock93}~\citeyear{leacock93}
used a word--context frequency matrix for WSD, in which each vector corresponds
to a type annotated with a sense tag. \citeauthor{yuret09}~\citeyear{yuret09} applied a
word--context frequency matrix to unsupervised WSD, achieving results
comparable to the performance of supervised WSD systems.

{\bf Context-sensitive spelling correction:} People frequently confuse certain
sets of words, such as {\em there}, {\em they're}, and {\em their}.
These confusions cannot be detected by a simple dictionary-based
spelling checker; they require context-sensitive spelling correction.
A word--context frequency matrix may be used to correct these kinds of
spelling errors \cite{jones97}.

{\bf Semantic role labeling:} The task of semantic role labeling
is to label parts of a sentence according to the roles they play
in the sentence, usually in terms of their connection to the main
verb of the sentence. \citeauthor{erk07}~\citeyear{erk07} presented a system in
which a word--context frequency matrix was used to improve the
performance of semantic role labeling. \citeauthor{pennacchiotti08}
\citeyear{pennacchiotti08} show that word--context matrices can reliably predict
the semantic frame to which an unknown lexical unit refers, with good levels of
accuracy. Such lexical unit induction is important in semantic role labeling, to
narrow the candidate set of roles of any observed lexical unit.

{\bf Query expansion:} Queries submitted to search engines such as Google and
Yahoo! often do not directly match the terms in the most relevant documents. To
alleviate this problem, the process of query expansion is used for generating
new search terms that are consistent with the intent of the original query. VSMs
form the basis of query semantics models \cite{cao08}. Some methods represent
queries by using session contexts, such as query cooccurrences in user sessions
\cite{huang03,jones06}, and others use click contexts, such as the urls that were
clicked on as a result of a query \cite{wen01}.

{\bf Textual advertising:} In pay-per-click advertising models, prevalent in search
engines such as Google and Yahoo!, users pay for keywords, called {\em bidterms}, which
are then used to display their ads when relevant queries are issued by users. The
scarcity of data makes ad matching difficult and, in response, several techniques for
bidterm expansion using VSMs have been proposed. The word--context matrix consists of
rows of bidterms and the columns (contexts) consist of advertiser identifiers
\cite{gleich04} or co-bidded bidterms (second order co-occurrences) \cite{chang09}.

{\bf Information extraction:} The field of information extraction (IE)
includes named entity recognition (NER: recognizing that a chunk
of text is the name of an entity, such as a person or a place), relation
extraction, event extraction, and fact extraction.
Pa{\c{s}}ca et al.~\citeyear{pasca06} demonstrate that a word--context frequency matrix
can facilitate fact extraction. \citeauthor{vyas09}~\citeyear{vyas09} propose a
semi-supervised model using a word--context matrix for building and iteratively
refining arbitrary classes of named entities.

\subsection{Pair--Pattern Matrices}
\label{subsec:pair--pattern-apps}

Pair--pattern matrices are most suited to measuring the semantic similarity of
word pairs and patterns (see Section~\ref{subsec:pair--pattern-vsms}).
For example, we can measure the similarity of two word pairs by the cosine of
the angle between their corresponding row vectors in a pair--pattern matrix.
There are many applications for measures of relational similarity.

{\bf Relational similarity:} Just as we can measure attributional similarity
by the cosine of the angle between row vectors in a word--context matrix,
we can measure relational similarity by the cosine of the angle between
rows in a pair--pattern matrix. This approach to measuring relational
similarity was introduced by \citeauthor{turney03b}~\citeyear{turney03b}
and examined in more detail by \citeauthor{turney05a}~\citeyear{turney05a}.
\citeauthor{turney06}~\citeyear{turney06} evaluated this approach to relational similarity
with 374 multiple-choice analogy questions from the SAT college entrance test,
achieving human-level performance (56\% correct for the pair--pattern matrix
and 57\% correct for the average US college applicant). This is the highest
performance so far for an algorithm. The best algorithm based on attributional
similarity has an accuracy of only 35\% \cite{turney06}. The best non-VSM algorithm
achieves 43\% \cite{veale04}.

{\bf Pattern similarity:} Instead of measuring the similarity between
row vectors in a pair--pattern matrix, we can measure the similarity between
columns; that is, we can measure pattern similarity. \citeauthor{lin01}
\citeyear{lin01} constructed a pair--pattern matrix in which the patterns
were derived from parsed text. Pattern similarity can be used to
infer that one phrase is a paraphrase of another phrase, which is useful
for natural language generation, text summarization, information retrieval,
and question answering.

{\bf Relational clustering:} Bi{\c{c}}ici and Yuret~\citeyear{bicici06} clustered
word pairs by representing them as row vectors in a pair--pattern matrix.
\citeauthor{davidov08}~\citeyear{davidov08} first clustered contexts
(patterns) and then identified representative pairs for each context cluster.
They used the representative pairs to automatically generate multiple-choice
analogy questions, in the style of SAT analogy questions.

{\bf Relational classification:} \citeauthor{chklovski04}~\citeyear{chklovski04}
used a pair--pattern matrix to classify pairs of verbs into semantic classes.
For example, {\em taint}$\,:\,${\em poison} is classified as {\em strength}
(poisoning is stronger than tainting) and {\em assess}$\,:\,${\em review}
is classified as {\em enablement} (assessing is enabled by reviewing).
\citeauthor{turney05b}~\citeyear{turney05b} used a pair--pattern matrix to classify noun
compounds into semantic classes. For example, {\em flu virus} is classified as
{\em cause} (the virus causes the flu), {\em home town} is classified as
{\em location} (the home is located in the town), and {\em weather report} is
classified as {\em topic} (the topic of the report is the weather).

{\bf Relational search:} \citeauthor{cafarella06}~\citeyear{cafarella06}
described relational search as the task of searching for entities that
satisfy given semantic relations. An example of a query for
a relational search engine is ``list all $X$ such that $X$ causes cancer''.
In this example, the relation, {\em cause}, and one of the terms in the
relation, {\em cancer}, are given by the user, and the task of the search engine
is to find terms that satisfy the user's query. The organizers of Task 4
in SemEval 2007 \cite{girju07} envisioned a two-step approach to relational
search: first a conventional search engine would look for candidate answers,
then a relational classification system would filter out incorrect answers.
The first step was manually simulated by the Task 4 organizers and the goal of
Task 4 was to design systems for the second step. This task attracted 14 teams
who submitted 15 systems. \citeauthor{nakov07}~\citeyear{nakov07}
achieved good results using a pair--pattern matrix.

{\bf Automatic thesaurus generation:} We discussed automatic thesaurus
generation in Section~\ref{subsec:word--context-apps}, with word--context
matrices, but arguably relational similarity is more relevant than
attributional similarity for thesaurus generation. For example, most
of the information in WordNet is in the relations between the words
rather than in the words individually. \citeauthor{snow06}~\citeyear{snow06}
used a pair--pattern matrix to build a hypernym-hyponym taxonomy, whereas
\citeauthor{pennacchiotti06}~\citeyear{pennacchiotti06} built a meronymy and causation
taxonomy. \citeauthor{turney08a}~\citeyear{turney08a} showed how a pair--pattern matrix can
distinguish synonyms from antonyms, synonyms from non-synonyms, and taxonomically similar
words ({\em hair} and {\em fur}) from words that are merely semantically
associated ({\em cradle} and {\em baby}).

{\bf Analogical mapping:} Proportional analogies have the form
$a\!:\!b\!::\!c\!:\!d$, which means ``$a$ is to $b$ as $c$ is to $d$''. For
example, {\em mason}$\,:\,${\em stone}$\,::\,${\em carpenter}$\,:\,${\em wood}
means ``mason is to stone as carpenter is to wood''. The 374 multiple-choice
analogy questions from the SAT college entrance test (mentioned above) all
involve proportional analogies. With a pair--pattern matrix, we can
solve proportional analogies by selecting the choice that maximizes
relational similarity (e.g., ${\rm sim_r}(mason\!:\!stone, carpenter\!:\!wood)$
has a high value). However, we often encounter analogies that involve
more than four terms. The well-known analogy between the solar system and the
Rutherford-Bohr model of the atom contains at least fourteen terms.
For the solar system, we have {\em planet, attracts, revolves, sun,
gravity, solar system}, and {\em mass}. For the atom, we have {\em revolves,
atom, attracts, electromagnetism, nucleus, charge}, and {\em electron}.
\citeauthor{turney08b}~\citeyear{turney08b} demonstrated that we can handle these
more complex, systematic analogies by decomposing them into sets
of proportional analogies.

\section{Alternative Approaches to Semantics}
\label{sec:alternatives}

The applications that we list in Section~\ref{sec:applications} do not
necessarily require a VSM approach. For each application, there are many
other possible approaches. In this section, we briefly consider a few of the main
alternatives.

Underlying the applications for term--document matrices
(Section~\ref{subsec:term--document-apps}) is the task of measuring the semantic
similarity of documents and queries. The main alternatives to VSMs for this task
are probabilistic models, such as the traditional probabilistic retrieval models
in information retrieval \cite{vanrijsbergen79,baezayates99} and the more recent
statistical language models inspired by information theory \cite{liu05}.
The idea of statistical language models for information retrieval is to
measure the similarity between a query and a document by creating a probabilistic
language model of the given document and then measuring the probability
of the given query according to the language model.

With progress in information retrieval, the distinction between the VSM
approach and the probabilistic approach is becoming blurred, as each approach
borrows ideas from the other. Language models typically involve multiplying
probabilities, but we can view this as adding logs of probabilities, which
makes some language models look similar to VSMs.

The applications for word--context matrices (Section~\ref{subsec:word--context-apps})
share the task of measuring the semantic similarity of words. The main
alternatives to VSMs for measuring word similarity are approaches that
use lexicons, such as WordNet \cite{resnik95,jiang97,hirst98,leacock98,budanitsky01}.
The idea is to view the lexicon as a graph, in which nodes correspond to
word senses and edges represent relations between words, such as hypernymy
and hyponymy. The similarity between two words is then proportional to the
length of the path in the graph that joins the two words.

Several approaches to measuring the semantic similarity of words combine
a VSM with a lexicon \cite{turney03b,pantel05,patwardhan06,mohammad06}. Humans
use both dictionary definitions and observations of word usage, so it is natural
to expect the best performance from algorithms that use both distributional and
lexical information.

Pair--pattern matrices (Section~\ref{subsec:pair--pattern-apps}) have in common
the task of measuring the semantic similarity of relations. As with
word--context matrices, the main alternatives are approaches that use
lexicons \cite{rosario01,rosario02,nastase03,veale03,veale04}.
The idea is to reduce relational similarity to attributional similarity,
${\rm sim_r}(a\!:\!b, c\!:\!d) \approx {\rm sim_a}(a,c) + {\rm sim_a}(b,d)$,
and then use a lexicon to measure attributional similarity. As we discuss
in Section~\ref{subsec:similarities}, this reduction does not work in
general. However, the reduction is often a good approximation, and there
is some evidence that a hybrid approach, combining a VSM with a lexicon,
can be beneficial \cite{turney03b,nastase06}.

\section{The Future of Vector Space Models of Semantics}
\label{sec:future}

Several authors have criticized VSMs \cite{french02b,pado03,morris04,budanitsky06}.
Most of the criticism stems from the fact that term--document and
word--context matrices typically ignore word order. In LSA, for instance, a phrase
is commonly represented by the sum of the vectors for the individual words in the
phrase; hence the phrases {\em house boat} and {\em boat house} will be represented
by the same vector, although they have different meanings. In English, word order
expresses relational information. Both {\em house boat} and {\em boat house} have a
Tool-Purpose relation, but {\em house boat} means Tool-Purpose$(boat, house)$
(a boat that serves as a house), whereas {\em boat house} means
Tool-Purpose$(house, boat)$ (a house for sheltering and storing boats).

\citeauthor{landauer02}~\citeyear{landauer02} estimates that 80\% of the meaning of English
text comes from word choice and the remaining 20\% comes from word order. However,
VSMs are not inherently limited to 80\% of the meaning of text.
\citeauthor{mitchell08}~\citeyear{mitchell08} propose composition models sensitive
to word order. For example, to make a simple additive model become syntax-aware,
they allow for different weightings of the contributions of the vector components.
Constituents that are more important to the composition therefore can participate
more actively. \citeauthor{clark07}~\citeyear{clark07} assigned distributional meaning
to sentences using the Hilbert space tensor product. \citeauthor{widdows08}~\citeyear{widdows08},
inspired by quantum mechanics, explores several operators for modeling composition
of meaning. Pair--pattern matrices are sensitive to the order of the words in a pair
\cite{turney06}. Thus there are several ways to handle word order in VSMs.

This raises the question, what are the limits of VSMs for semantics?
Can all semantics be represented with VSMs? There is much that we do not yet
know how to represent with VSMs. For example, \citeauthor{widdows04}~\citeyear{widdows04}
and \citeauthor{rijsbergen04}~\citeyear{rijsbergen04} show how disjunction, conjunction,
and negation can be represented with vectors, but we do not yet know how
to represent arbitrary statements in first-order predicate calculus. However,
it seems possible that future work may discover answers to these limitations.

In this survey, we have assumed that VSMs are composed of elements with
values that are derived from event frequencies. This ties VSMs to
some form of distributional hypothesis (see Sections \ref{subsec:vector-motivation}
and \ref{subsec:hypotheses}); therefore the limits of VSMs depend on the limits
of the family of distributional hypotheses. Are statistical patterns of word
usage sufficient to figure out what people mean? This is arguably the major
open question of VSMs, and the answer will determine the future of VSMs.
We do not have a strong argument one way or the other, but we believe that the
continuing progress with VSMs suggests we are far from reaching their limits.

\section{Conclusions}
\label{sec:conclusions}

When we want information or help from a person, we use words
to make a request or describe a problem, and the person replies with
words. Unfortunately, computers do not understand human language, so
we are forced to use artificial languages and unnatural user interfaces.
In science fiction, we dream of computers that understand human language,
that can listen to us and talk with us. To achieve the full potential
of computers, we must enable them to understand the semantics of natural
language. VSMs are likely to be part of the solution to the problem
of computing semantics.

Many researchers who have struggled with the problem of semantics
have come to the conclusion that the meaning of words is closely
connected to the statistics of word usage (Section~\ref{subsec:hypotheses}).
When we try to make this intuition precise, we soon find we are
working with vectors of values derived from event frequencies; that is,
we are dealing with VSMs.

In this survey, we have organized past work with VSMs according to
the structure of the matrix (term--document, word--context, or pair--pattern).
We believe that the structure of the matrix is the most important
factor in determining the types of applications that are possible.
The linguistic processing (Section~\ref{sec:linguistic}) and
mathematical processing (Section~\ref{sec:mathematical}) play
smaller (but important) roles.

Our goal in this survey has been to show the breadth and power
of VSMs, to introduce VSMs to those who less familiar with them,
and to provide a new perspective on VSMs to those who are already
familiar with them. We hope that our emphasis on the structure of
the matrix will inspire new research. There is no reason to believe
that the three matrix types we present here exhaust the possibilities.
We expect new matrix types and new tensors will open up more
applications for VSMs. It seems possible to us that all of the
semantics of human language might one day be captured in some
kind of VSM.

\acks{Thanks to Annie Zaenen for prompting this paper. Thanks to
Saif Mohammad and Mariana Soffer for their
comments. Thanks to Arkady Borkovsky and Eric Crestan for developing the
distributed sparse-matrix multiplication algorithm, and to Marco
Pennacchiotti for his invaluable comments. Thanks to the anonymous
reviewers of {\em JAIR} for their very helpful comments and
suggestions.}

\vskip 0.2in

\bibliography{turney10a}
\bibliographystyle{theapa}

\end{document}